\documentclass[lettersize,journal]{IEEEtran}
\usepackage{amsmath,amsfonts}
\usepackage{algorithmic}
\usepackage{algorithm}
\usepackage{array}
\usepackage{textcomp}
\usepackage{stfloats}
\usepackage{subfigure}
\usepackage{url}
\usepackage{verbatim}
\usepackage{graphicx}
\usepackage{cite}
\usepackage[utf8]{inputenc} \usepackage{amssymb} \usepackage{multirow} \usepackage{graphicx} \usepackage{tabularx} \usepackage[acronym]{glossaries} \usepackage{makecell} \usepackage{rotating} \usepackage{threeparttable} \usepackage{booktabs} \usepackage{ragged2e} \usepackage{array}
\usepackage{cite}
\usepackage{epstopdf}
\usepackage{epsfig}
\usepackage{bm}
\usepackage{mathrsfs}
\usepackage{amsfonts}
\usepackage{amssymb}
\usepackage{amsmath}
\usepackage{latexsym}
\usepackage{graphicx}
\usepackage{subcaption}
\usepackage{amsthm}
\usepackage{bbding}
\usepackage{hyperref}
\usepackage{indentfirst}
\usepackage{algorithm} 
\usepackage{algorithmic} 
\usepackage{multirow} 
\usepackage{xcolor}
\usepackage{color}
\usepackage{longtable}
\DeclareUnicodeCharacter{2002}{}
\usepackage{graphics}
\usepackage{graphicx}
\usepackage{epsfig}
\usepackage[switch]{lineno}
\usepackage{hyperref}

\usepackage{hyperref}
\usepackage[hyphenbreaks]{breakurl}

\usepackage{stfloats}

\theoremstyle{plain}

\newtheorem{myDef}{Definition}

\theoremstyle{definition}

\theoremstyle{remark}

\begin{document}

\title{Employing Iterative Feature Selection in Fuzzy Rule-Based Binary Classification}
\author{Haoning~Li,
Cong Wang,~\IEEEmembership{Member,~IEEE},
and Qinghua Huang

\thanks{This work was supported in part by National Key Research and Development Program under Grant 2018AAA0102100, in part by National Natural Science Foundation of China under Grants 62206220, 62071382 and 82030047, in part by Guangdong Basic and Applied Basic Research Foundation under Grant No. 2021A1515110019, in part by Young Talent Fund of Association for Science and Technology in Shaanxi, China, in part by Fundamental Research Funds for the Central Universities, and in part by Innovation Capability Support Program of Shaanxi under Grant 2021TD-57. (Corresponding author: Qinghua Huang). Haoning Li and Cong Wang are contributed equally to this work.}
\thanks{Haoning Li is with the Shool of Artificial Intelligence, Optics and Electronics, Northwestern Polytechnical University, Xi'an 710072, China (e-mail: 1134067491@qq.com).}
\thanks{Cong Wang is with the Shool of Artificial Intelligence, Optics and Electronics, Northwestern Polytechnical University, Xi'an 710072, China and also with the Research \& Development Institute of Northwestern Polytechnical University in Shenzhen, Shenzhen 51800, P.R. China (e-mail: congwang0705@nwpu.edu.cn).}
\thanks{Qinghua Huang is with the Shool of Artificial Intelligence, Optics and Electronics, Northwestern Polytechnical University, Xi'an 710072, China (e-mail: qhhuang@nwpu.edu.cn).}
}

\maketitle

\begin{abstract}
The feature selection in a traditional binary classification algorithm is always used in the stage of dataset preprocessing, which makes the obtained features not necessarily the best ones for the classification algorithm, thus affecting the classification performance. For a traditional rule-based binary classification algorithm, classification rules are usually deterministic, which results in the fuzzy information contained in the rules being ignored. To do so, this paper employs iterative feature selection in fuzzy rule-based binary classification. The proposed algorithm combines feature selection based on fuzzy correlation family with rule mining based on biclustering. It first conducts biclustering on the dataset after feature selection. Then it conducts feature selection again for the biclusters according to the feedback of biclusters evaluation. In this way, an iterative feature selection framework is build. During the iteration process, it stops until the obtained bicluster meets the requirements. In addition, the rule membership function is introduced to extract vectorized fuzzy rules from the bicluster and construct weak classifiers. The weak classifiers with good classification performance are selected by Adaptive Boosting and the strong classifier is constructed by ``weighted average''. Finally, we perform the proposed algorithm on different datasets and compare it with other peers. Experimental results show that it achieves good classification performance and outperforms its peers.
\end{abstract}

\begin{IEEEkeywords}
Fuzzy rule, iterative feature selection, feedback of bicluster, rule membership function.
\end{IEEEkeywords}
\section{Introduction}
\IEEEPARstart{C}{lassification} problems play an important role in machine learning. A multi-classification problem can be seen as a combination of multiple binary classification problems. For now, such problems have appeared widely in numerous practical areas, such as image recognition \cite{ref1}, medical diagnosis \cite{ref2}, and text analysis \cite{ref3}. Many researchers constantly try to develop better methods to deal with them.

Traditional binary classification methods include K-nearestneighbor (KNN) \cite{ref4}, support vector machine (SVM) \cite{ref5}, decision tree \cite{ref6}, and Bayesian classification \cite{ref7}. To this day, they have been used in various fields to solve binary classification problems and have achieved good performance. However, with the diversification of current binary classification tasks, researchers are no longer satisfied with just getting the results of binary classification. In fact, they want to know more about the process of classification. The classification algorithms should be able to express the classification process more concretely and vividly so that people can understand it better \cite{ref8}. At this time, rule-based classification methods have gradually attracted the extensive attention of researchers. Differing from traditional classification methods, they can extract classification rules from datasets and show those rules concretely. Specifically, the rules are represented in the form of natural language or vectors. Therefore, practitioners can also understand the process of classification. In general, it is divided into two parts: 1) mining rules from a large number of examples and 2) classifying objects according to the rules \cite{ref9}.

Biclustering is a common classification method based on rule mining \cite{ref10}. Through it, a large number of biclusters are obtained, where each bicluster contains a large amount of information related to classification. By extracting rules from biclusters and then integrating them, we can get classifiers composed of such rules. Compared with its peers, biclustering has a more interpretable classification process. In the classification process, the meaningful expression patterns are first extracted from data matrix by biclustering, the extracted biclusters contain classification information. Then, specific classification rules are extracted from biclusters by using classification rule extraction methods and represent them as vectors, and finally classify them.

For SVM, KNN, decision tree, Bayesian classification or rule-based biclustering classification, each of them has its corresponding limitations. When they are applied to datasets of different types and sizes, their performance may deteriorate. If multiple classifiers can be integrated to classify the dataset and complement each other, the classification performance will be greatly improved. To realize this idea, researchers have begun to try to use ensemble learning to solve binary classification problems and have achieved good results \cite{ref11,ref12,ref13}.

In order to solve more complex binary classification problems, binary classification algorithms are gradually applied to large sample and attributes datasets. Such a dataset usually contains a large number of redundant features or ones that have no impact on the sample category. The existence of these features may make the running process of the algorithm more complex so as to increase its computational complexity. To make matters worse, it may lead to the decline of the performance of the binary classification algorithm thus resulting in unsatisfactory results \cite{ref14}. In order to effectively shorten the running time of the classification algorithm and get better classification results, researchers usually use feature selection methods to eliminate redundant or irrelevant features. The combination of feature selection and classification algorithms has been widely used in practical fields, see \cite{ref15}, \cite{ref16}. In particular, attribute reduction based on fuzzy rough set is an effective and commonly used feature selection method.

When combined with the binary classification algorithm, it is usually used independently of the binary classification algorithm. The original dataset is processed by employing a feature selection. Redundant and irrelevant features in the dataset are removed to obtain a new dataset. The new dataset is trained and classified by using the binary classification algorithm. In this way, in addition to partial redundancy and irrelevant features to a certain extent, the computational load is reduced and the performance of the binary classification algorithm is increased. However, the feature selection and the training process of the binary classification algorithm are not related. After feature selection is made by a certain method, various binary classification methods are trained according to the data set after feature selection. Since the feature selection method is not interacted with the binary classification algorithm, we do not know the requirements of the binary classification algorithm for feature selection. Such feature selection is blind and incomplete and often does not meet the requirements of different binary classification methods. Therefore, how to combine the feature selection process with the classification process to eliminate their independence, so as to obtain the corresponding optimal feature selection results according to different classification algorithms has become an important problem.

As we know, the binary classification algorithm based on biclustering is a rule-based method. Researchers use the biclustering algorithm to search for biclusters. They extract classification rules from the biclusters. Finally, the dataset is classified based on the classification rules. In the previous rules extracted from bicluster, for convenience, researchers usually define rules as deterministic rules. That is, a rule consisting of a specific set of features in a bicluster corresponds to only one class of samples. However, in fact, the sample types corresponding to the same set of feature values in the bicluster may be different. In the binary classification problem, it may occur that the rules extracted by biclustering correspond to two classes of samples at the same time. Therefore, the extracted rules are not deterministic rules, those are actually uncertain rules, which can also be called fuzzy rules. Therefore, how to convert the deterministic rules in the rule-based binary classification algorithm into fuzzy rules becomes a difficult problem. Whereas, to introduce fuzzy rules, fuzzy theory is essential and many researchers are working on it \cite{ref17,ref18,ref19}. As a basis for the study of this problem, some work about fuzzy theory and fuzzy clustering have been carried out in our previous studies \cite{ref20,ref21,ref22,ref23}.

\begin{figure}[!h]
\centering
\includegraphics[width=1\linewidth]{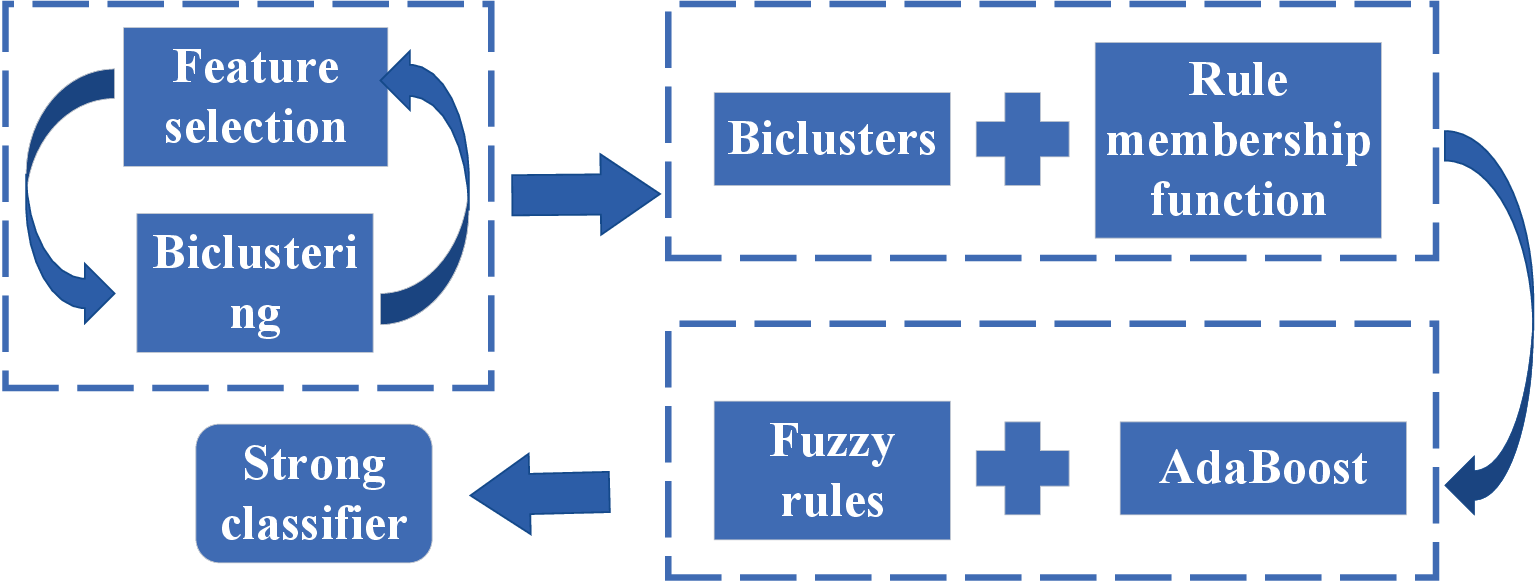}
\caption{Architecture of the proposed algorithm.}
\label{figure1}
\end{figure}

The current binary classification methods mentioned above have two problems: 1) feature selection and classification algorithms are fragmented and do not have interactivity; 2) the mined rules are not fuzzy rules. In order to solve these problems, we propose a novel fuzzy rule-based binary classification algorithm with an iterative feature selection framework. In it, we combine a feature selection algorithm based on fuzzy correlation family with a biclustering algorithm based on heuristic search to build an iterative feature selection framework, as shown in Fig. \ref{figure1}. First, feature selection is carried out by an attribute reduction algorithm based on fuzzy correlation family, then bicluster search is carried out by a heuristic method. Second, feature selection is carried out again according to the feedback of biclustering results, then bicluster search is carried out again by the heuristic method, until the obtained biclusters meet the requirements. In this way, an iterative feature selection framework is constructed. Third, after obtaining the required biclusters, fuzzy rules are extracted from the biclusters by defining and introducing the rule membership function. The fuzzy rules are classified according to the membership degree and  weak classifiers are constructed. Finally, weak classifiers with good performance are selected through an ensemble learning algorithm based on AdaBoost and a strong classifier is constructed according to ``weighted average''. By using the strong classifier to classify the dataset, the final classification results are obtained.

The main contributions of this work are described as follows:
\begin{itemize}
\item An iterative feature selection framework is proposed, which combines feature selection based on fuzzy rough sets with biclustering. In the framework, feature selection is performed based on feedback from biclustering results. Both feature selection and biclustering are performed at the same time, which promote each other and work together to ultimately obtain better biclusters.

\item A rule membership function is introduced to convert the original precise rules into fuzzy rules when mining rules in biclusters. Compared to precise rules, fuzzy rules can retain more information in biclusters and improve the accuracy of rule-based classification.

\item The rule membership function is introduced into ensemble learning. A weak classifier set is first constructed based on the fuzzy rules extracted from biclusters, then some better performing weak classifiers are selected from the set by AdaBoost method. After that, the rule membership function is introduced into the weighted voting process, the weak classifiers are aggregated into a strong classifier as the final classifier.
\end{itemize}

Compared with other binary classification methods, the proposed algorithm has the following innovations:
\begin{itemize}
\item It constructs an iterative feature selection framework that combines the process of feature selection with the process of searching for biclusters. In traditional methods, feature selection and classification algorithms are independent of each other and there is no information interaction between them. Therefore, the features selected by feature selection are not necessarily the best features that meet the requirements of current classification algorithms. After increasing the interactivity between them, feature selection can selectively iterate based on the feedback from the biclustering results, enabling better features to be selected and better biclusters to be obtained.
\item In traditional rule based binary classification algorithms, classification rules mined from data sets are often converted into deterministic rules for convenience. This process can lead to the loss of useful information of classification, ultimately resulting in poor results in rule-based classification. In this work, by introducing a rule membership function, the fuzziness of the rules is preserved and the loss of information is avoided. The resulting rules have better classification capabilities.
\end{itemize}

Section II introduces related work about binary classification. After that, we give details of the proposed algorithm in Section III. Then we present the experiments results and analysis in Section IV. Finally, Section V concludes this paper.

\section{Related Work}\label{sec:methodology}

Since the binary classification problem has become an important issue in the field of machine learning, many researchers have devoted themselves to researching and applying better algorithms to solve this problem. This section describes the related work of different types of methods to binary classification problems.

\subsection{Traditional Machine Learning Methods}

Traditional machine learning methods such as KNN, SVM, decision tree, and Bayes have been widely applied to various binary classification problems due to their simple structure and convenient use. To now, there still are many binary classification algorithms based on these methods that have been continuously proposed. In order to be applied to the current field, researchers have combined many current technologies with these traditional methods to further optimize and improve the performance of these methods. For example, Das and Jena \cite{ref24} use the combination of  local binary pattern (LBP) and completed local binary pattern (GLRLM) features along KNN for texture classification. Yang \textit{et al.}\cite{ref25} use SVM algorithm to classify the data and image of haematococcus pluvialis. Sardari and Eftekhari \cite{ref26} design a fuzzy decision tree for imbalanced data classification. Sunori \textit{et al.}\cite{ref27} use Baysian method to estimate rainfall an classify the level of rainfall. Although these methods have been continuously optimized and their performance is getting better, they still have the problem of unclear classification processes, which is difficult to understand.

\subsection{Rule-Based Biclustering Methods}
Biclustering, as a method for extracting information from data, has been used since it was proposed. In 2000, Cheng and Church \cite{ref28} first proposed the concept of biclustering and a new method to search for biclusters by using a mean-square-residue score function, which is used in studying biological gene expression data. After that, many researchers devote themselves to the optimization and improvement of biclustering algorithms, thus yielding many variants such as OPSM, QUBIC, and BICPAM.

As the classification problems become increasingly complex, in order to better meet the requirements of mining consistency rules through biclustering, researchers introduce evolutionary algorithms to biclustering to solve classification problems. Bleuler \textit{et al.} \cite{ref29} firstly describe the architecture and implementation of a general EA framework in biclustering. On this basis, researchers have continuously improved the evolutionary biclustering method. For example, Divina and Aguilar-Tuiz \cite{ref30} propose an EA biclustering algorithm using a new single fitness function, which considers four different objectives. Huang \textit{et al.} \cite{ref31} propose a new evolutionary strategy based on bi-phase to search for biclusters and optimize the search space, which has a significant improvement on the efficiency and effectiveness of biclustering. The development of evolutionary biclustering make mining classification rules faster and better.

\subsection{Ensemble Learning Methods}
In fact, ensemble learning-based decision-making methods have appeared since the beginning of civilized society. For example, citizens vote to select officials or make laws. Dietterich \textit{et al.} \cite{ref32} explain three basic reasons for the success of ensemble learning from a mathematical perspective: statistical, computational and representational. In 1979, Dasarathy and Sheela \cite{ref33} came up with the idea of ensemble learning for the first time. After that, researchers have studied and proposed more efficient ensemble learning methods, such as Boosting \cite{ref34}, AdaBoost \cite{ref35}, Bagging \cite{ref36}, and Random Forest \cite{ref37}. Among them, AdaBoost and Random Forest are mostly used and have the best performance.

Ensemble learning integrates multiple basic classifiers together, comprehensively evaluating the classification results of multiple classifiers and obtaining the final classification results. In ensemble learning, each classifier can complement each other, which eliminates the limitations of a single classifier. Therefore, the classification results obtained through integration have a higher accuracy. In practical applications, ensemble learning is often combined with other classification methods as an algorithm for binary classification problems. For example, Liu and Zhou \cite{ref38} propose an ensemble learning method for COVID-19 fact verification. Huang \textit{et al.} \cite{ref39,ref40,ref41,ref42,ref43,ref44,ref45,ref46,ref47,ref48} have done extensive research on breast ultrasound images and combine AdaBoost and biclustering for breast tumor classification \cite{ref49}.

\subsection{Feature Selection Methods Based on Fuzzy Correlation}

When making feature selection, it is often necessary to measure the correlation among features.  Pearson's correlation coefficient, which is most often used for variable correlation calculations, is no longer applicable when the feature variables contain a large amount of uncertain information. Therefore, in order to solve the problem of calculating the correlation among features containing uncertain information, the concept of fuzzy correlation has been applied to numerous feature selection methods. Eftekhari \textit{et al.} \cite{ref50} propose a distributed feature selection method based on the concept of hesitant fuzzy correlation, which can effectively realize feature selection for high-dimensional microarray datasets. Bhuyan \textit{et al.} \cite{ref51} use correlation coefficients and fuzzy models to select features and sub-features for classification. Ejegwa \textit{et al.} \cite{ref52} apply some modified Pythagorean fuzzy correlation measures in determining certain selected decision problems. Numerous fuzzy correlation based algorithms have been proposed and a large portion of them have been used to perform feature selection of datasets in various domains. In addition, these methods are combined with various classification algorithms to solve the problems existing in the field of classification.

\subsection{Feature Selection Methods Based on Fuzzy Rough Sets}

In 1982, Pawlak et al. proposed the concept of rough set, which is a mathematical tool that can quantitatively analyze and process imprecise, inconsistent and incomplete information. When rough set characterizes uncertain information, its purpose is to derive the classification rules of the problem through attribute reduction while keeping the classification ability unchanged. However, the classification by rough set is based on the equivalence relations, which limits the application of the rough set. The discretization of numerical data causes information loss, therefore, rough set is not suitable for numerical data. In order to solve this problem, researchers extend equivalence relations to coverings, binary relations, and neighborhood systems. After covering rough set is proposed, the attribute reduction method based on traditional rough set is also extended to covering rough set. The most common method is to use the discernibility matrix to reduce attributes. Discernibility matrix is proposed by Skowron and Rauszer, which is a classical tool to reduce attributes \cite{ref54}. Unfortunately, only a few of cover rough sets are suitable for attribute reduction using discernibility matrix method. Therefore, Yang \textit{et al.} \cite{ref55} propose a new approach based on related family to compute the attribute reducts of other covering rough sets.

Although rough set is generalized, such as covering rough set, it still has considerable limitations in solving real-value datasets. Therefore, fuzzy set is introduced into rough set theory to improve the performance in solving the real-value datasets. Fuzzy set was proposed by Zadeh \cite{ref56} in 1965, which is a mathematical tool for studying uncertain problems. It starts with the fuzziness of knowledge, emphasizes the ambiguity of set boundary, and depicts the fuzziness of things through membership function. The rough set introduced by fuzzy theory is called fuzzy rough set, which is applied in various fields \cite{ref57,ref58}. Fuzzy set theory has also been introduced to cover rough set to solve the real-value data problem. Accordingly, researchers have working on attribute reduction algorithm based on fuzzy covering rough set. The attribute reduction method  based on relate family proposed by Yang  is introduced into fuzzy covering rough set and improve into fuzzy relate family method. This method is used by Yang in attribute reduction of fuzzy rough covering set, which is combined with decision tree method to diagnose thyroid diseases. In addition, many scholars have proposed and applied the attribute reduction algorithm based on fuzzy rough set and its extension, and combined it with classifiers such as SVM, decision tree, KNN, and biclustering to select features and classify datasets, which achieves good results.

Inspired by the above work, we propose a fuzzy rule-based binary classification method that integrates feature selection, biclustering, and ensemble learning. The proposed algorithm implements classification tasks based on classification rules, which are extracted from biclusters. The iterative feature selection process is based on the feedback of biclusters, so it directly affects the biclusters, which also affects the classification rules extracted from the biclusters, and ultimately affects the classification algorithm. 

\section{Proposed Methodology}
\subsection{Problem Description}
Let us assume that a dataset, as a data matrix $M$, has $R$ rows and $C$ columns. Its bicluster is a submatrix $N$ of $M$, where $N$ is composed of the row subset and column subset of $M$, respectively. In general, rows represent samples and columns represent features. In the binary classification problem, some feature subsets in the bicluster are universal, and some samples are consistent under such subsets. Then these samples can be divided into same class and these feature subsets are the corresponding classification rules.

In the real binary classification problem, we often use feature extraction methods to extract the data matrix of samples and features. However, there will be a large number of features in the data matrix. Some of them are useless in classification, which are redundant or completely irrelevant to the sample category. When using the binary classification algorithm to classify the data matrix, all features are generally put into the calculation. At this time, redundant or irrelevant features greatly increase the computational overhead of the binary classification algorithm. To solve this problem, researchers have introduced and developed feature selection methods. In this way, the obtained features are screened and selected again to remove feature redundancy. However, the traditional feature selection methods are generally used as a preprocessing step of the classification algorithm. As a result, feature selection and classification are separate from each other. We cannot verify whether the result of feature selection is good or not according to the feedback of binary classification results, and whether the feature subset obtained by feature selection can meet the requirements of classification. In addition, different binary classification algorithms may use different feature subsets when classifying the same datasets. Hence, feature selection methods should be targeted. Different feature subsets should also be used for classification according to different binary classification algorithms.

In addition to the limitations of feature selection, rule-based binary classification algorithms have several problems in rule extraction. The main purpose of rule extraction is to obtain deterministic rules. For example, Huang et al. only select the category that accounts for more than $65\%$ of the biclusters when using biclusters for rule extraction, and use this category as the rule category extracted from the biclusters. Although this proposed method is simple and convenient, it ignores another class of samples corresponding to the feature combination (i.e., rule) in the bicluster. Therefore, such deterministic rules have an inherent error rate, which may cause large error when classification. This may lead to the decline of classification performance and difficulty in optimization.

\subsection{Model and Algorithm}

In order to solve the above problems in feature selection and rule extraction, we propose a fuzzy rule-based binary classification algorithm with iterative feature selection. Obviously, the proposed algorithm is divided into three parts: 1) iterative feature selection framework consisting of feature selection of fuzzy correlation family and biclustering based on a heuristic algorithm, 2) fuzzy rule extraction based on fuzzy membership function, and 3) ensemble learning based on the AdaBoost. The specific idea of the proposed algorithm is presented as follows: the original dataset is first selected by using a fuzzy correlation family algorithm based on fuzzy rough sets. The feature selected dataset is searched by a heuristic biclustering algorithm to mine classification rules. By judging the support degree $S$ of biclusters, we verify whether the biclusters meets our requirements. If not, the feature selection of the fuzzy correlation family will be conducted again for the bicluster and then the bicluster search will be conducted again. Just like this, by keep iterators, the iterative feature selection framework formed. Rule mining and feature selection are carried out at the same time. They promote each other until the optimal biclusters are obtained. Next, the rule membership function is introduced to extract the fuzzy rules from the biclusters. The fuzzy rules are classified and the weak classifiers are constructed according to the rule membership function. Finally, the weak classifiers are tested by AdaBoost to screen out the ones with good performance and then the strong classifier is constructed by weighted voting.

\begin{figure}[!h]
\centering
\includegraphics[width=1\linewidth]{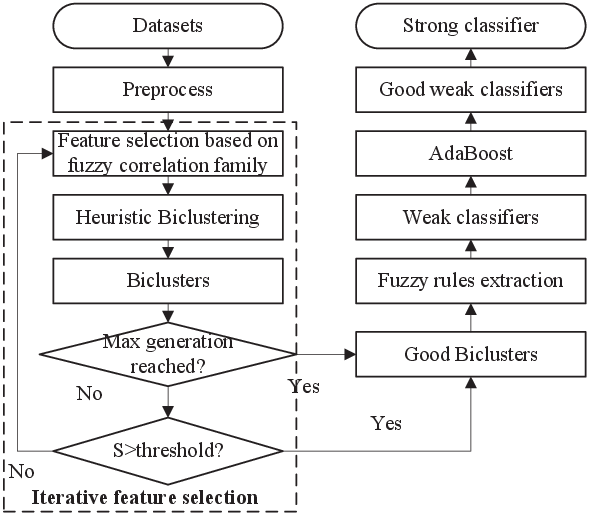}
\caption{Flow chat of the proposed algorithm.}
\label{figure2}
\end{figure}

\subsection{Iterative Feature Selection Framework}
\subsubsection{Feature selection based on fuzzy correlation family attribute reduction}
For a traditional rough set, its data structure is discrete data matrix. Therefore, many attribute reduction methods, such as discernibility matrix, information entropy and dependency can be applied to each data set and achieve good results. However, when discrete data are extended to continuous ones, rough sets are also extended to fuzzy fashions. The above methods have great limitations when applied to continuous data. Therefore, attribute reduction methods based on fuzzy correlation families are proposed and can be well applied to fuzzy rough sets.

Suppose that $U$ is a nonempty finite universe. Each sample can be described by a condition attribute set $A$ and a symbolic decision attribute set $D=\{d_1,d_2,\cdots,d_n\}$, in the case of $A\cap D$, $(U, A\cap D)$ is called a fuzzy decision table. For each condition attribute $a\in A$, a fuzzy binary relation $R_a$ is defined. For any $x,y\in U$, if $R_a$ satisfies reflexivity $(R(x,y)=1)$, symmetry $(R(x,y)=R(y,x))$ and transitivity $(R(x,y)\leq \sup\limits_{z\in U}\min(R(x,y),R(y,x))$, then $R_a$ is called a fuzzy equivalence relation. A set $B\subseteq A$ is defined as a fuzzy equivalence relation, which can be expressed as $R_B=\mathop{\cap}\limits_{a\in B}R_a$ .

We assume that $F(U)$ is the set of all fuzziness defined on $U$ and $B$ is subset of $A$. For $x\in U$, a pair of upper and lower approximation operators can be defined as:

\begin{align}
\underline{R_B}F(x)=\inf\limits_{u\in U}\max\{1-R_B(x,u),F(u)\}
\end{align}
\begin{align}
\overline{R_B}F(x)=\sup\limits_{u\in U}\min\{R_B(x,u),F(u)\}
\end{align}
where $\underline{R_B}(F)(x)$ means that degree $x$ is completely subordinate to $F$, while $\overline{R_B}(F)(x)$ implies that degree $x$ may be subordinate to $F$. $(\underline{R_B}(F)(x),\overline{R_B}(F)(x))$ is called the fuzzy rough set of $F$ about $B$. For $(U,A\cup D)$ and $B\subseteq A$, the fuzzy positive field of $D$ with respect to $B$ is defined as:

\begin{align}
POS_B(D)=\mathop{\cup}\limits_{F\in U/D}\underline{R_B}(F)
\end{align}

For $x\in U$, there exists $POS_B(D)(x)=\underline{R_B}([F]_D)(x)$. By keeping the positive field unchanged, the attribute reduction of fuzzy rough set is defined as follows:

\begin{myDef}
Assume that $(U, A\cup D)$ is a fuzzy decision table. If attribute subset $A$ meets two conditions: 1) for $x\in U$, we have $\underline{R_P}([x]_D)(x)\leq \underline{R_A}([x]_D)(x)$ and 2) for $a\in P$, there exists $y\in U$, $R_{P-\{a\}}([y]_D)(y)\le \underline{R_A}([y]_D)(y)$, then the attribute subset $P\subseteq A$ is called the reduction of $A$ relative to $D$.
\end{myDef}

Given a dataset $T=(U,A,D)$, where $U=\{x_1,x_2, \cdots, \\ x_n\}$ is a sample set, $A=\{a_1,a_2,\cdots,a_m\}$ is a condition attribute set, and $D=\{d_1,d_2,\cdots,d_t\}$ is the decision attribute set. For each $x_j\in U$, $a_i(x_j)\in A$ indicates that sample $x_j$ belongs to attribute $a_i$. The particles generated by taking with the sample $x_j$ as the center point are called fuzzy neighborhood relations. The particle set generated by each sample forms a fuzzy coverage. In this paper, according to different particle radius $\delta \in [0,1]$, a fuzzy covering element $A_{\delta,g_r}$ is generated from a given center point $g_r\in G$. Its fuzzy membership is defined as:

\begin{equation}
A_{\delta,g_r}(x_i)=\begin{cases}
\frac{\delta-|a_i(x_j)-g_r|}{\delta},&|a_i(x_j)-g_r|\leq\delta\\
0,&|a_i(x_j)-g_r|>\delta
\end{cases}
\end{equation}

\noindent where $g_r$ represents the particle center point, that is, it forms a fixed point coverage for the center point $g_r$. $G$ is a set collecting the particle center points on $[0,1]$. The value range starts from $0$ and increases with equal difference of particle radius. When the last particle exceeds $1$, the particle center point is taken as $1$. By taking $\delta=0.3$ as an example, the particles in $[0,1]$ are denoted as $G=\{0,0.3,0.6,0.9,1\}$. Each fuzzy coverage corresponds to a condition attribute and the fuzzy coverages generated by all the condition attributes form a family of fuzzy coverages. In this way, the dataset $T=(U,A,D)$ is transformed into a fuzzy coverage information system $\widetilde{S}=(U,\hat{\Delta},D)$. In a fuzzy $\beta$ coverage, fuzzy neighborhood $N(x)=\{x_j|d(x_i,x_j)\geq(1-\beta)*e\}$ can be determined by the size of neighborhood radius $e$. If $d(x_i,x_j)=\delta$, then $A(x)=0$; if $d(x_i,x_j)=0$, then $A(x)=1$. This ensures that all sample objects can be divided into different decision classes. The attribute reduction algorithm based on fuzzy correlation family is divided into two parts: 1) the original data is standardized to calculate the fuzzy correlation family and 2) the attribute reduction subset is obtained based on the fuzzy correlation family.

\subsubsection{Biclustering based on heuristic algorithm}
After analyzing the dataset selected by fuzzy correlation family features, it can be seen that the samples have consistency on some attribute columns. The consistency information is the classification rule that we would like to extract. Here the biclusters extracted from the binary dataset belong to the ``column consistency" type of biclusters. Therefore, a more appropriate entropy-based evaluation method, Mean Entropy Score (MES), is used in this paper. It is defined as

\begin{equation}
{\rm MES}=\frac{1}{|C|}\sum\limits_{j\in C}e_j
\end{equation}

\begin{equation}
e_j=-\mathop{\sum}\limits_{i\in N_{p}}\frac{p(i)}{n}\log_2\frac{p_i}{n}=\log_2n-\frac{1}{n}\sum\limits_{i\in N_{p}}p_i\log_2 p_i
\end{equation}

\noindent where $e_j$ represents the information entropy of $j$th column. The higher the entropy is, the more uncertain information it contains, and the greater the uncertainty of the biclusters composed of such columns. $N_{p}$ is the population number of hierarchical clustering in this column, $p_i$ is the number of group members corresponding to population $i$ in the hierarchical clustering in this column, and $n$ is the number of samples (rows) contained in the submatrix. 

The information entropy of a column represents the clutter level of the data cluster group of the column. The more clusters corresponding to the column, the more dispersed the column values are, and the greater the corresponding entropy is. If a column of values belong to the same cluster, the entropy value is equal to $0$. Therefore, the lower the average entropy score MES of each column, the better the quality of the biclustering corresponding to the submatrix. The biclustering algorithm is divided into three steps: 1) extracting bicluster seeds, 2) heuristic searching for bicluster based on bicluster seeds, and 3) subsequent integration of bicluster.

In the process of bicluster seed extraction, we use the method based on agglomerative hierarchical clustering to cluster each column of the data matrix. The clustering results of each column may be a part of the final biclustering, thus we take the clustering results of each column as the starting point. We then realize heuristic search for biclustering. As a result, the clustering results of each column become the biclustering seeds.

After obtaining a series of bicluster seeds, we conduct heuristic search on each seed to build a bicluster and use MES as the quality evaluation function to conduct bicluster evaluation. The biclustering process is summarized in Algorithm \ref{algorithm2}. It should be noted that in steps 10 and 11 of the algorithm, the deletion of rows or columns is not done randomly, but to ensure that MES of data matrix after deleting the rows or columns is minimized.

\begin{algorithm}[htb]
\caption{Overall procedure of biclustering}
{\bf Input:} Distance formula between classes $dist(c_i,c_j)$, preset termination merge threshold $t$, preset termination threshold $\delta$, and number of sample objects $N$.\\
{\bf Output:} Biclusters.
\begin{algorithmic}[1]
\STATE Each object is regard as a spearate class, and the Euclidean distance between objects is used to initilize the distance matrix $D$. Assuming the minimum inter class distance between classes is $dist_{min}$.
\REPEAT
\STATE Find the two classes $c_i$ and $c_j$ that minimize $dist(c_i,c_j)$.\\
\STATE If $dist(c_i,c_j) < t$, both $c_i$ and $c_j$ are merged into one class.\\
\STATE Use the distance calculation function $dist(c_i,c_j)$ to update the distance matrix $D$.
\UNTIL {$dist_{min} > t$}
\STATE The initial matrix $N=(R,C)$ with full columns is obtained based on biclustering seeds.\\
\STATE Calculate the $\rm{MES}$ of original matrix.
\REPEAT
\STATE Calculate the $\rm{MES}$ of the new matrix after removing a row or column.
\STATE Remove the row or column corresponding to the $\rm{MES}$ from the original matrix.
\UNTIL {$\rm{MES} < \delta$}
\STATE Obtain biclusters based on the biclustering seeds after the above steps.
\end{algorithmic}
\label{algorithm2}
\end{algorithm}

\subsubsection{Construction of iterative feature selection framework}

The iterative feature selection framework is mainly divided into two parts: 1) feature selection algorithm and 2) classification rule mining. For feature selection, we use attribute reduction based on fuzzy correlation family. For classification rule mining, we use a biclustering algorithm based on a heuristic algorithm to search for biclusters. Then we define and introduce the support degree of bicluster to evaluate the bicluster and select features again according to the feedback of evaluation results. In this way, the iteration continues until the bicluster support degree meets the requirements. The support degree $S$ is defined as:

\begin{align}
S=\frac{N_{\max}}{N},N_{\max}=\max\{N_A,N_B\}
\end{align}

\noindent where $N_A$ and $N_B$ respectively represent the number of samples of two categories included in the bicluster, and the support degree $S$ of the bicluster represents the number of sample categories with the largest number in the bicluster accounting for the total number of samples in the bicluster. The higher the support degree of the bicluster, the closer to 1, which means that the classification rules extracted from the bicluster have stronger classification ability. The closer to 0.5, the worse the classification ability is, and more likely to be random guesses. In this algorithm, we take the support degree $S$ as the evaluation index of the bicluster in the iterative feature selection process. When the bicluster obtained by the heuristic algorithm does not meet the preset threshold of $S$, we reselect the feature and research biclusters and continue to iterate until the given conditions are met or the maximum number of iterations is reached.

\subsection{Extraction of Fuzzy Rules}

As a local correlation pattern, biclusters can reveal certain classification rules. If all features of a bicluster have the same value, and most of the samples belong to the same category, the bicluster represents a certain classification rule of the samples. For the sake of simplicity, we convert the rules contained in the bicluster into vectorization ones. The rule extraction process from the bicluster is expressed as:

\begin{equation}
R=[r_1,r_2,\cdots, r_t]
\end{equation}

\noindent where $r_i$ represents the $i$th feature in the pattern rule, namely the feature value of the $i$th column in the bicluster matrix.

\begin{figure}[!h]
\centering
\includegraphics[width=1\linewidth]{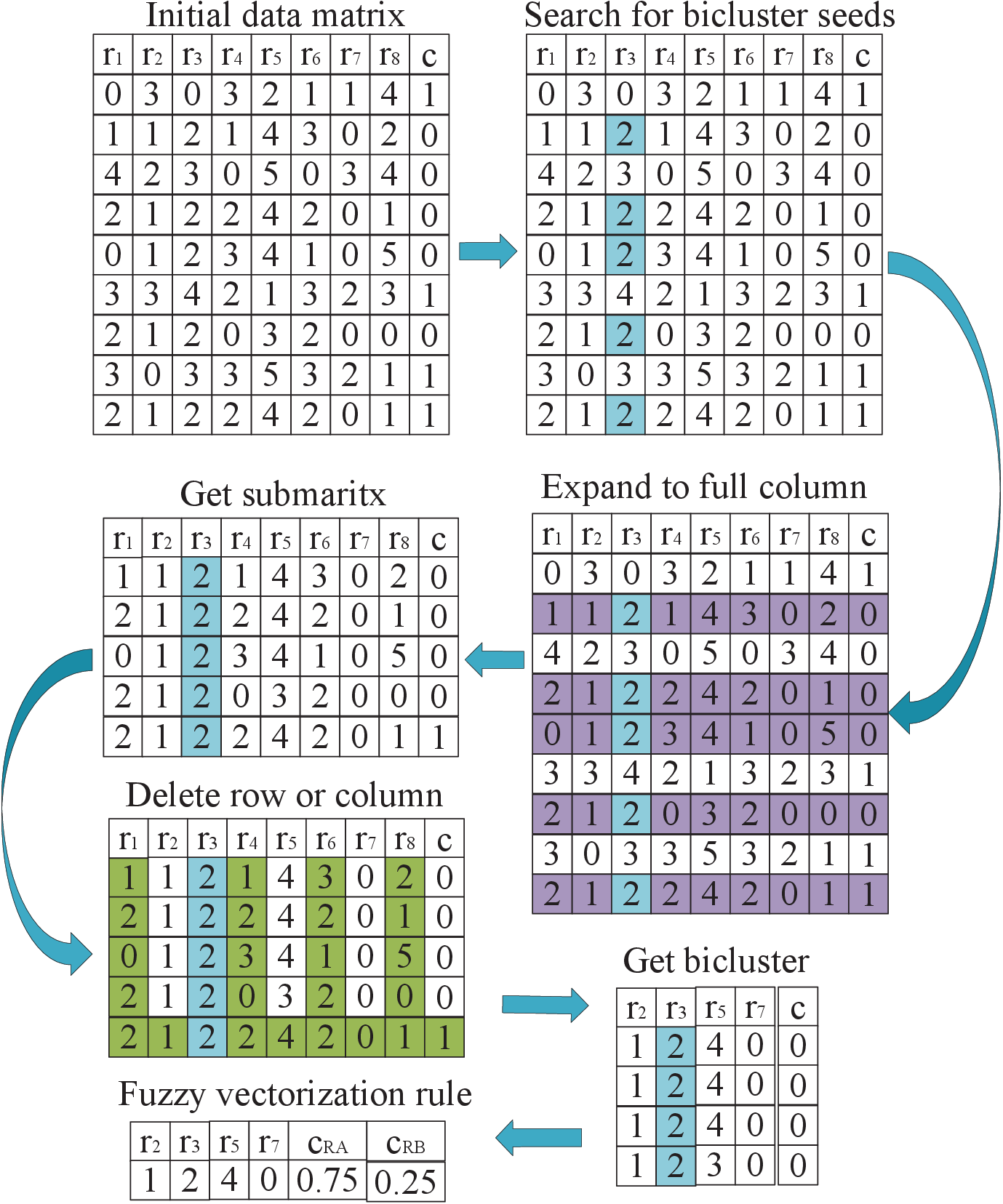}
\caption{Process of extracting fuzzy rules. }
\label{figure3}
\end{figure}

Since the samples corresponding to the same features in biclustering are not of a single category, the extracted rules are not deterministic rules, but fuzzy rules. Therefore, we introduce a rule membership function to describe the process of rule extraction:

\begin{align}
\sum_{i=1}^{k}C_{Ri}=1,C_{Ri}=\frac{N_i}{N},C_{Ri}\in[0,1]
\end{align}

\noindent where $R$ represents the vectorization rule extracted from the bicluster and $k$ represents the category count of the sample. $C_{Ri}$ represents the membership of rule $R$ to the $i$th category. The sum of membership of rule $R$ to all categories is 1. Each rule has its corresponding membership degree of each category, which represents the probability that samples meeting this rule belong to each category. It is determined by the proportion of the number of samples of each category in the total number of samples in the bicluster. $N_i$ represents the number (rows) of samples with the $i$th category in the bicluster and $N$ is the total number of samples in the bicluster. The process of extracting fuzzy rules is shown in Fig. \ref{figure3}.

\subsection{Ensemble Learning based on AdaBoost Algorithm}
Since the classification rules extracted from each bicluster belong to local features, a single classifier composed of each rule cannot well describe the features of the entire dataset. As a result, the classifier does not perform satisfactorily in the test set and does not have the ability to generalize, which performs well in the training set. Therefore, we introduce the AdaBoost algorithm to integrate weak classifiers into strong ones with high accuracy and strong generalization ability.

When determining rule attributes, we denote the membership of rule $R$ to two categories $A$ and $B$ as $C_{RA}$ and $C_{RB}$. The rule belongs to the category with high degree of membership, which can be defined as:

\begin{equation}
P_R=\begin{cases}
A,&C_{RA}> C_{RB},\\
B,&C_{RA}> \leq C_{RB}.
\end{cases}
\end{equation}
\noindent where $P_R$ represents the $R$ rule category, and the fuzzy rules extracted from all the biclusters are divided into two categories: $A$ and $B$.

\begin{figure}[htbp]
\centering
\includegraphics[width=1\linewidth]{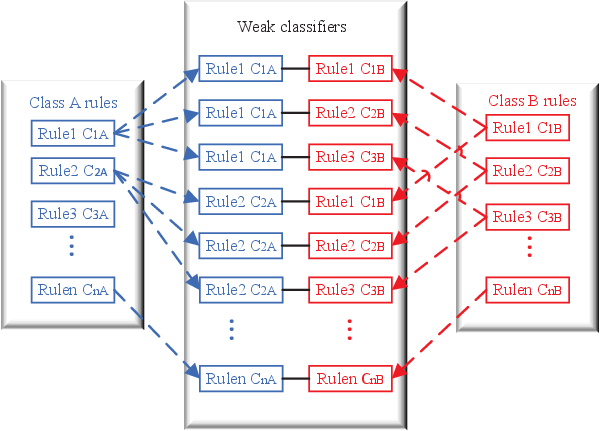}
\caption{Construction of weak classification based on fuzzy rules.}
\label{figure4}
\end{figure}

After the categories of the rules are determined and classified, weak classifiers for AdaBoost training are built. Note that, two rules are took out from two categories of rule sets to construct a weak classifier. The construction process is shown in Fig. \ref{figure4}.

As shown in Fig. \ref{figure4}, although the rules are divided into two categories in the process of building a weak classifier, the rule membership degree of both categories are still stored in the weak classifier. The rule membership degree is used in the weight calculation later. For example, in the first weak classifier, $C_{1A}$ indicates the degree to which the first rule of class $A$ that constitutes the classifier belongs to $A$ and $C_{1B}$ indicates the degree to which the first rule of class $B$ that constitutes the classifier belongs to $B$.

After the weak classifiers are constructed, AdaBoost iterative algorithm is used to learn a strong classifier with high accuracy and good generalization performance. As an iterative algorithm, the core idea of AdaBoost is to train multiple individual learners for different weight distributions of a unified training sample set, and then give the final decision output in a ``weighted average'' way. The whole process of AdaBoost is defined as follows:

Suppose a binary classified dataset is $D=(x_i, y_i): i=1, \\ 2, \cdots, m$, where $x_i$ is the input feature and $y_i$ is the classified label.

\textbf{Step 1}: Initialize the weight value of samples:

\begin{equation}
w^{1}_{i}=\frac{1}{N}, i=1,2,\cdots,N
\end{equation}

\textbf{Step 2}: Iterate the weak classifiers according to the following steps:

\begin{itemize}
\item Train weak classifier $h_{t}(x)$, where $x$ is the input sample.

\item Calculate the weight value of the current weak classifier $\alpha_{t}$, which is decided by the error rate of the weak classifier on the training set.

\begin{equation}
\epsilon_t=\sum_{i=1}^N w_i^{(t)} \cdot I\left(h_t\left(\mathbf{x}_i\right) \neq y_i\right)
\end{equation}

\begin{equation}
\alpha_t=\frac{1}{2} \ln \left(\frac{1-\epsilon_t}{\epsilon_t}\right)
\end{equation}

\noindent where $I$ is the indicator function.

\item For the next iteration, update sample weights.

\begin{equation}
w_i^{(t+1)}=\frac{w_i^{(t)} \cdot \exp \left(-\alpha_t \cdot y_i \cdot h_t\left(\mathbf{x}_i\right)\right)}{Z_t}
\end{equation}

\begin{equation}
Z_t=\sum_{i=1}^N w_i^{(t)} \cdot \exp \left(-\alpha_t \cdot y_i \cdot h_t\left(\mathbf{x}_i\right)\right)
\end{equation}

\noindent where $Z_t$ is the normalization factor.

\end{itemize}

\textbf{Step 3}: Obtain the final strong classifier that is defined as

\begin{equation}
H(\mathbf{x})=\operatorname{sign}\left(\sum_{t=1}^T \alpha_t \cdot h_t(\mathbf{x})\right)
\end{equation}

\begin{figure}[htbp]
\centering
\includegraphics[width=1\linewidth]{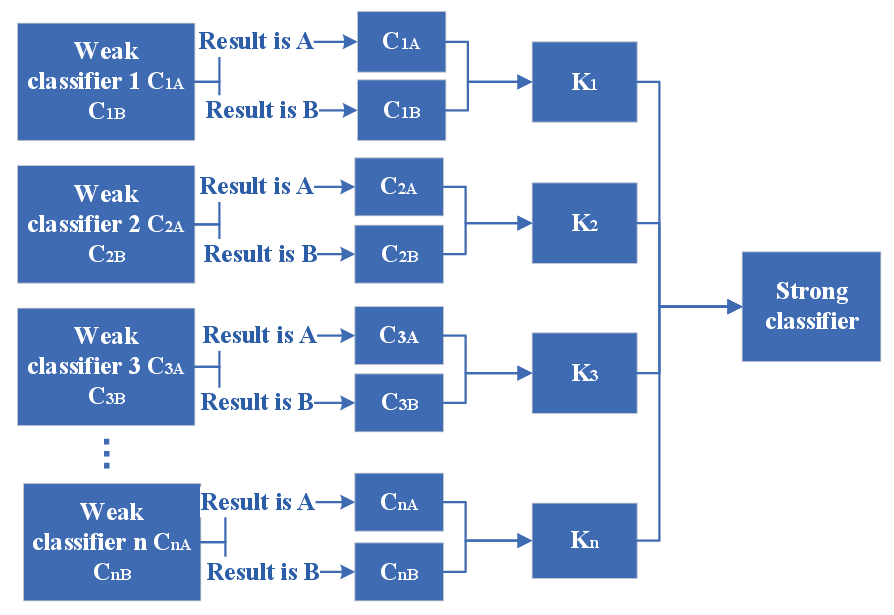}
\caption{Weighted average process based on AdaBoost.}
\label{figure5}
\end{figure}

After the effective weak classifier set is formed, the weak classification in the set is used for ``weighted average'' combination to form a strong classifier. The ``weight'' here includes not only the weight of the weak classifier given after the iteration, but also the membership of the rules in the weak classifier. Weighted average process based on AdaBoost is shown in Fig. \ref{figure5}.

When using the weak classifiers to form a strong classifier for weighted voting, the $n$th weak classifier classifies a sample. If the classification result is $A$, its weight when participating in voting is $C_{nA}\times K_n$, if the classification result is $B$, its weight when participating in voting is $C_{nB}\times K_n$. In this way, each weak classifier can participate in voting during classification, thus obtaining the final classification result of the strong classifier. In addition, differing from the traditional weight calculation of AdaBoost, we introduce the rule membership of fuzzy rules into the weight calculation, which further improves the precision of the obtained strong classifier.

\section{Experimental Study}

\subsection{Comparative Experiments}
In order to verify the classification performance of the proposed algorithm, we design a set of comparative experiments. The proposed algorithm and its three peers, abbreviated as HCL \cite{ref49}, YLL \cite{ref55}, and SHD \cite{ref59}, are respectively applied to eight real datasets for classification experiments. Experimental results prove the superiority of the proposed algorithm. These eight datasets are named as Waveform, Spambase, Sonar, Clean, Wdbc, Pima, Ionosphere, and Divorce, which are selected from public commonly used as binary classification datasets in UCI. In addition, we choose six evaluation indicators, i.e., Accuracy, Precision, Recall, Specificity, Receiver Operating Characteristic (ROC) curve, and Area Under Curve (AUC). The calculation formulas of these first four indicators are expressed as:
\begin{align}
\rm{Accuracy}=\frac{TP+TN}{TP+FN+FP+TN}
\end{align}
\begin{align}
\rm{Precision}=\frac{TP}{TP+FP}
\end{align}
\begin{align}
\rm{Recall}=\frac{TP}{TP+FN}
\end{align}
\begin{align}
\rm{Specificity}=\frac{TN}{TN+FP}
\end{align}
where more details are presented as follows:
\begin{itemize}
\item True Positive (TP): both the predicted value and the real value are positive.
\item False Positive (FP): the predicted value is positive and the true value is negative.
\item True Negative (TN): both the predicted value and the true value are negative.
\item False Negative (FN): the predicted value is negative and the true value is positive.
\end{itemize}

In addition, ROC represents a curve drawn with pseudo normal class rate as abscissa and true class rate as ordinate. Generally, the closer the curve is to the upper left corner, that is, the larger the area under the curve, the better the classifier performance. The advantage of ROC as an evaluation index is that it is not affected by sample imbalance. No matter how to change the proportion of two types of samples in the test set, ROC remains basically unchanged. In order to accurately describe ROC, we use the area under AUC calculation curve as the evaluation index.

In addition, we adopt a ten fold cross validation method in the comparative test, which is a commonly used test method in the experiment. During the experiment on each data set, we divide the data set into ten parts, and take turns to use nine of them as the training set and one as the test set for the experiment. Each experiment will obtain the corresponding accuracy, precision, recall and other test results. We use the average of the ten results as the final result to evaluate.

Before implementing the experiment, we need to set some parameters in the algorithm. The threshold for bicluster support degree is empirically set to 0.65. The maximum number of iterations is empirically set to 100. In addition, we need to determine $\delta$, $\beta$, and $G(g_r \in G)$, where $\delta$ represents the radius of the neighborhood, $\beta$ represents the threshold of the fuzzy intercept, and $g_r$ denotes the number of coverage elements uniformly distributed in $[0,1]$. For experimental testing, we first take parameter $\delta$ and vary it in steps of 0.05, however, the final performance is not good when $\delta \in [0.5,1]$. This is since a larger value of $\delta$ indicates a larger neighborhood radius, and when it approaches 1, generating fuzzy coverings loses its meaning. Therefore, we specify the ideal interval of values for $\delta$ as $[0,0.5]$ and vary it in steps of 0.02. In the course of extensive experiments, we finalize the value of parameter $\beta=0.5$. The value of $g_r$ can be adjusted as the number of samples increases. For example, when there are 7000 samples in the dataset, we take $g_r=80$.

\begin{table*}
    \centering
    \caption{\textsc{Results on eight datasets}}
\label{experiment}
    \begin{tabular}{lcccccccccccc}
        \toprule
        & \multicolumn{4}{c}{Waveform} & \multicolumn{4}{c}{Spambase} & \multicolumn{4}{c}{Sonar}\\
        \midrule
        Algorithm & Accuracy & Precision & Recall & Specificity & Accuracy & Precision & Recall & Specificity & Accuracy & Precision & Recall & Specificity  \\
        \midrule
        HCL & 0.899 & 0.848 & 0.977 & 0.817 & 0.918 & 0.892 & 0.978 & 0.832 & 0.907 & 0.880 & 0.974 & 0.807\\
        YLL & 0.910 & 0.919 & 0.897 & 0.924 & 0.911 & 0.900 & 0.951 & 0.858 & 0.926 & 0.925 & 0.953 & 0.884\\
        SHD & 0.916 & 0.909 & 0.920 & 0.913 & 0.924 & 0.920 & 0.939 & 0.907 & 0.928 & 0.918 & 0.963 & 0.877\\
        Ours & 0.940 & 0.948 & 0.943 & 0.938 & 0.950 & 0.950 & 0.967 & 0.926 & 0.935 & 0.938 & 0.945 & 0.923\\
        \midrule
        & \multicolumn{4}{c}{Clean} & \multicolumn{4}{c}{Wdbc} & \multicolumn{4}{c}{Pima}\\
        \midrule
        Algorithm & Accuracy & Precision & Recall & Specificity & Accuracy & Precision & Recall & Specificity & Accuracy & Precision & Recall & Specificity  \\
        \midrule
        HCL & 0.888 & 0.907 & 0.907 & 0.860 & 0.965 & 0.956 & 0.991 & 0.919 & 0.741 & 0.781 & 0.893 & 0.344\\
        YLL & 0.900 & 0.750 & 1.000 & 0.857 & 0.918 & 0.979 & 0.885 & 0.970 & 0.629 & 0.671 & 0.775 & 0.400\\
        SHD & 0.869 & 0.864 & 0.864 & 0.857 & 0.953 & 0.963 & 0.963 & 0.935 & 0.779 & 0.802 & 0.872 & 0.610\\
        Ours & 0.958 & 0.928 & 1.000 & 0.910 & 0.971 & 1.000 & 1.000 & 0.979 & 0.795 & 0.621 & 0.692 & 0.836\\
        \midrule
        & \multicolumn{4}{c}{Ionosphere} & \multicolumn{4}{c}{Divorce}\\
        \midrule
        Algorithm & Accuracy & Precision & Recall & Specificity & Accuracy & Precision & Recall & Specificity \\
        \midrule
        HCL & 0.821 & 0.835 & 0.630 & 0.967 & 0.980 & 0.964 & 1.000 & 0.958 \\
        YLL & 0.906 & 0.900 & 0.692 & 0.975 & 0.961 & 0.923 & 1.000 & 0.926 \\
        SHD & 0.833 & 0.833 & 0.714 & 0.909 & 0.961 & 0.963 & 0.963 & 0.958 \\
        Ours & 0.930 & 0.878 & 1.000 & 0.857 & 1.000 & 1.000 & 1.000 & 1.000 \\
        \bottomrule
    \end{tabular}
\end{table*}

From Table \ref{experiment}, it can be seen that the proposed algorithm performs well on almost all the metrics for each dataset compared to other three algorithms. Specifically speaking, on Waveform dataset, the proposed algorithm improves 0.032, 0.056, 0.034, and 0.017 on the four metrics of Accuracy, Precision, Recall, and Specificity, respectively, compared to the average of the other three algorithms. On Clean dataset, the proposed algorithm improves an average of 0.073, 0.088, 0.031, and 0.052. On Spambase dataset, the proposed algorithm improves on average by 0.032, 0.030, 0.011, and 0.009. On Sonar dataset, the proposed algorithm improves on average by 0.015, 0.016, 0.017, and 0.016. On Wdbc dataset, the proposed algorithm improves on average by 0.026, 0.028, 0.020, and 0.020. On Pima dataset, the proposed algorithm improves on average by 0.079, 0.076, -0.079, and 0.023. On Inosphere dataset, the proposed algorithm improves on average by 0.077, 0.070, 0.102, and 0.078. On Divorce dataset, the proposed algorithm improves by 0.033, 0.027, 0.012, and 0.019 on average.

In most of the datasets, the proposed algorithm achieves relatively good results in terms of Accuracy, Precision, Recall, and Specificity, especially in Waveform, Clean, Spambase, Sonar, and Ionosphere datasets. In Wdbc and Divorce datasets, the proposed algorithm achieves very high values for all metrics, especially for Recall, where it reaches a perfect 1.000. The exception is Pima dataset, where the proposed algorithm outperforms the other methods in Precision and Specificity, but has a slightly lower Recall value.

\begin{figure}
  \centering
  \subfigure[]{\includegraphics[width=0.45\linewidth,height=0.278\linewidth]{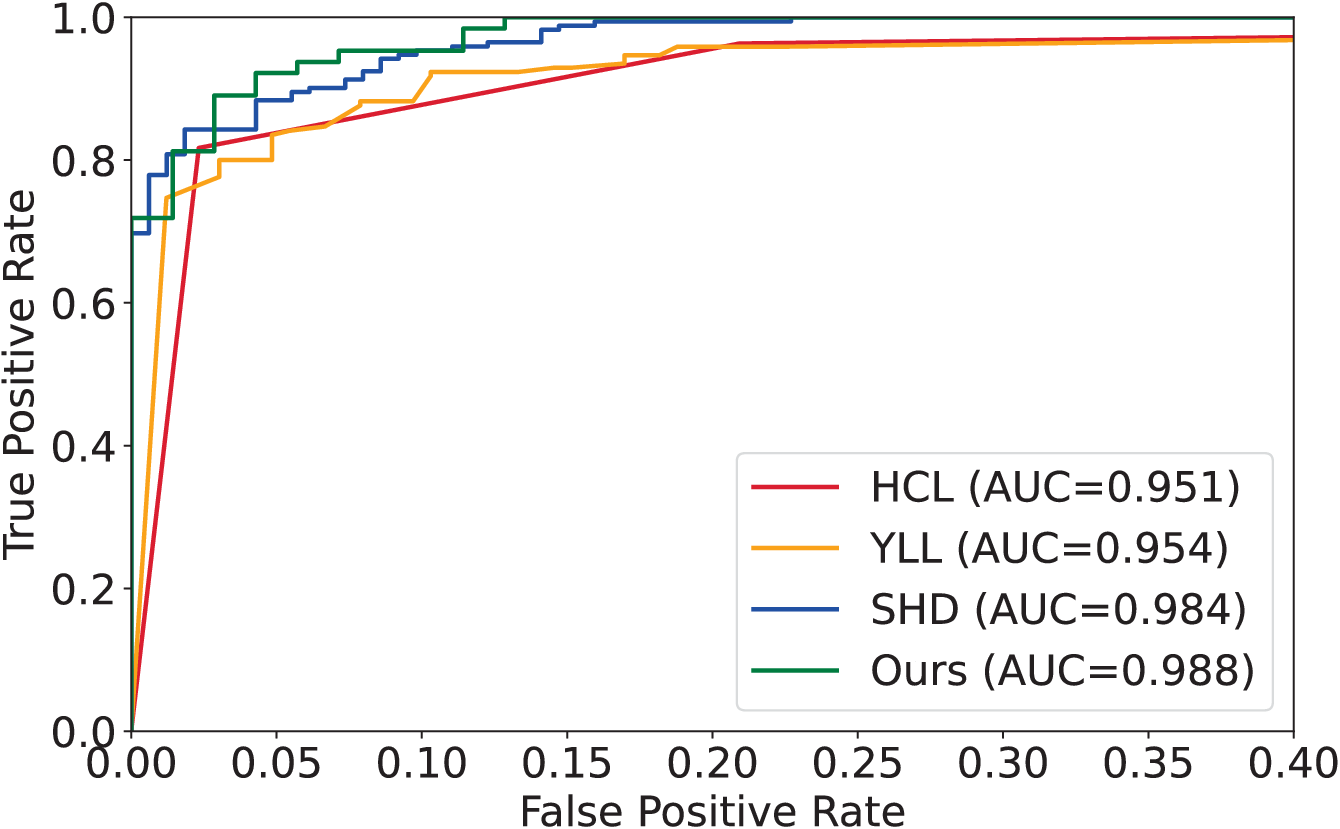}}
  \subfigure[]{\includegraphics[width=0.45\linewidth,height=0.278\linewidth]{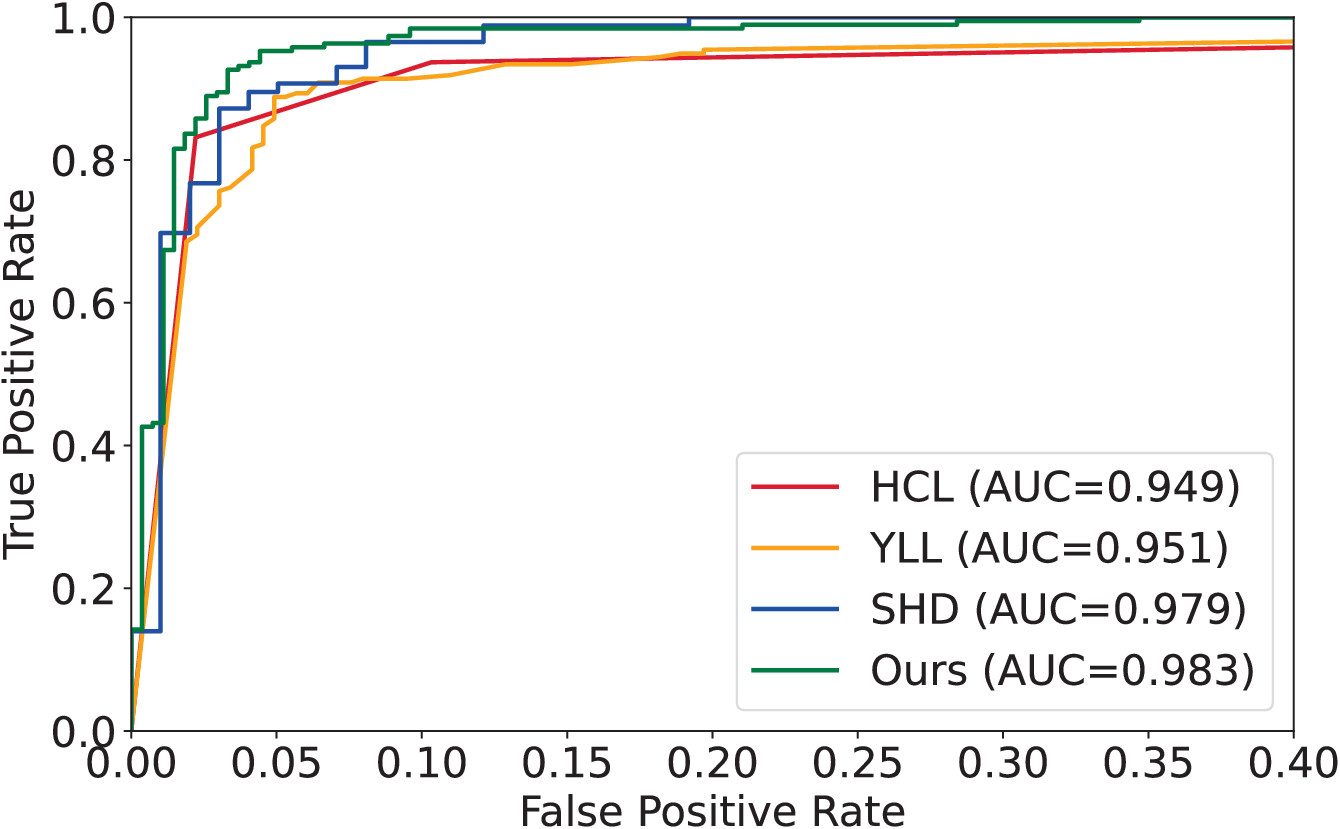}}

  \subfigure[]{\includegraphics[width=0.45\linewidth,height=0.278\linewidth]{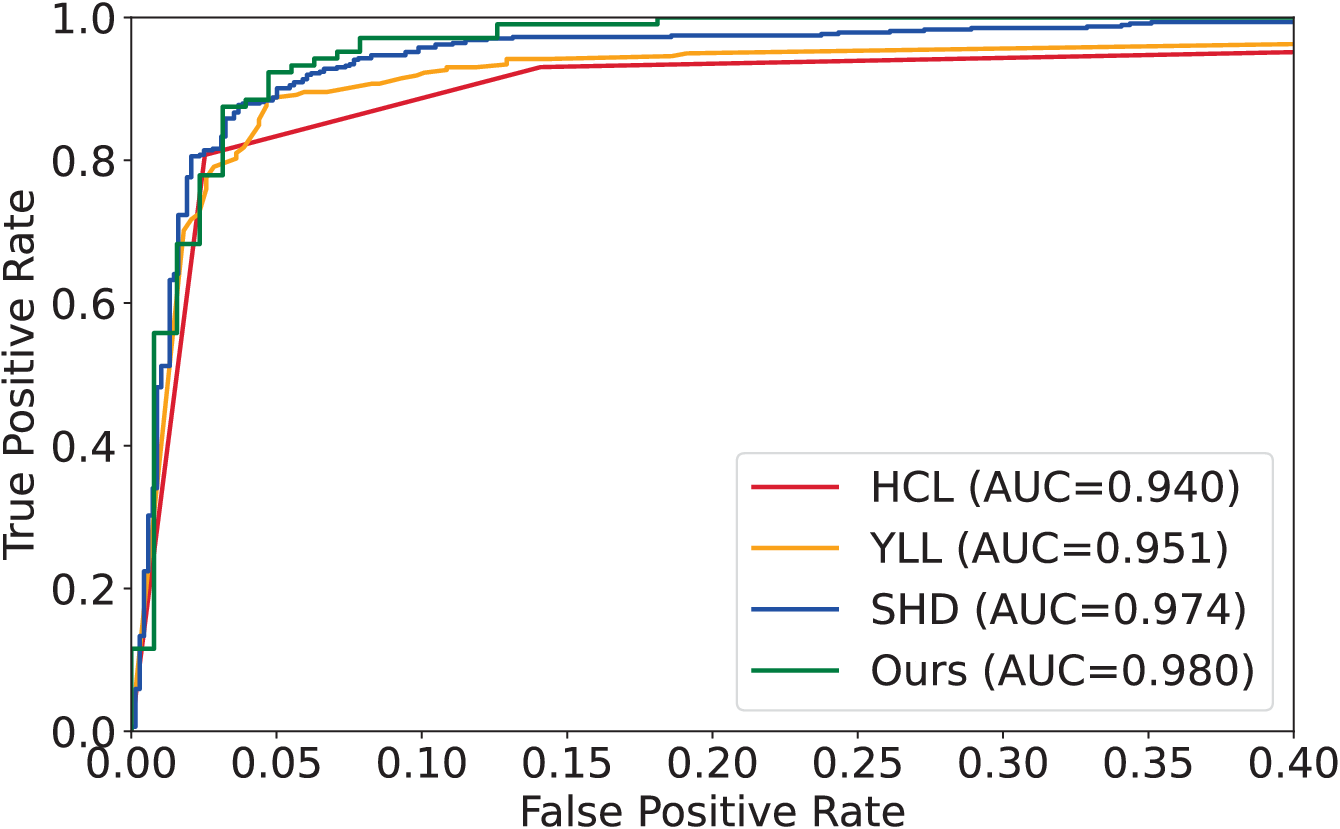}}
  \subfigure[]{\includegraphics[width=0.45\linewidth,height=0.278\linewidth]{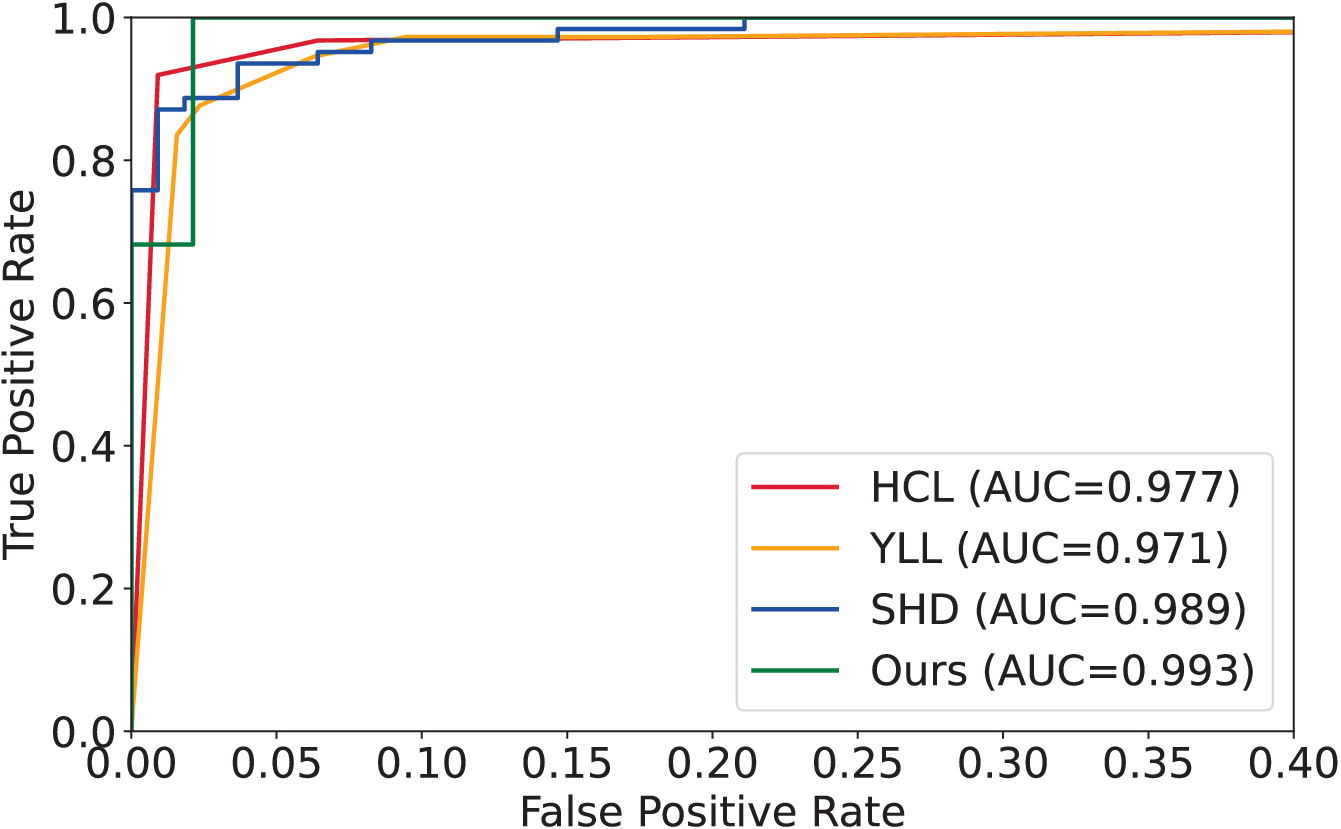}}

  \subfigure[]{\includegraphics[width=0.45\linewidth,height=0.278\linewidth]{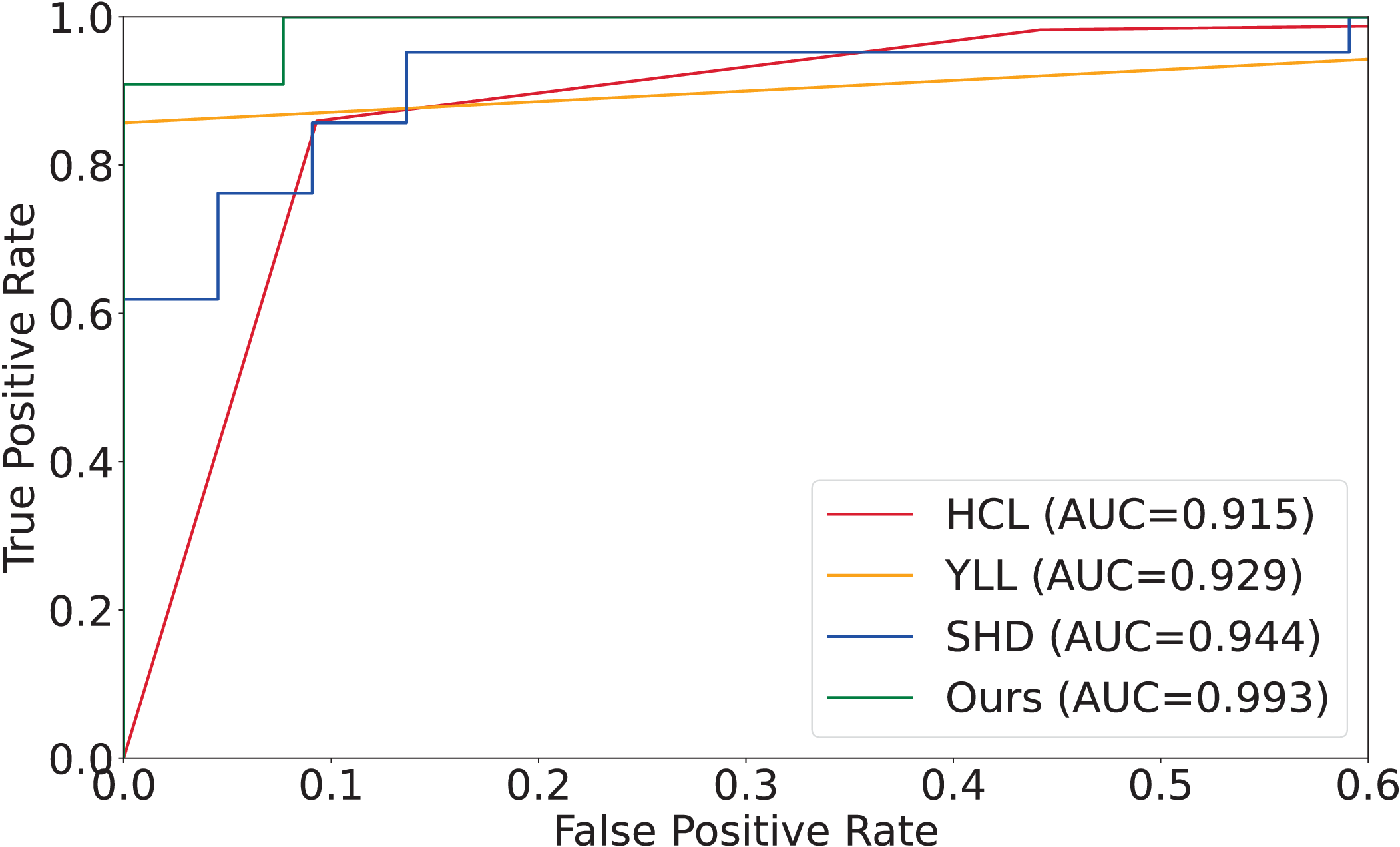}}
  \subfigure[]{\includegraphics[width=0.45\linewidth,height=0.278\linewidth]{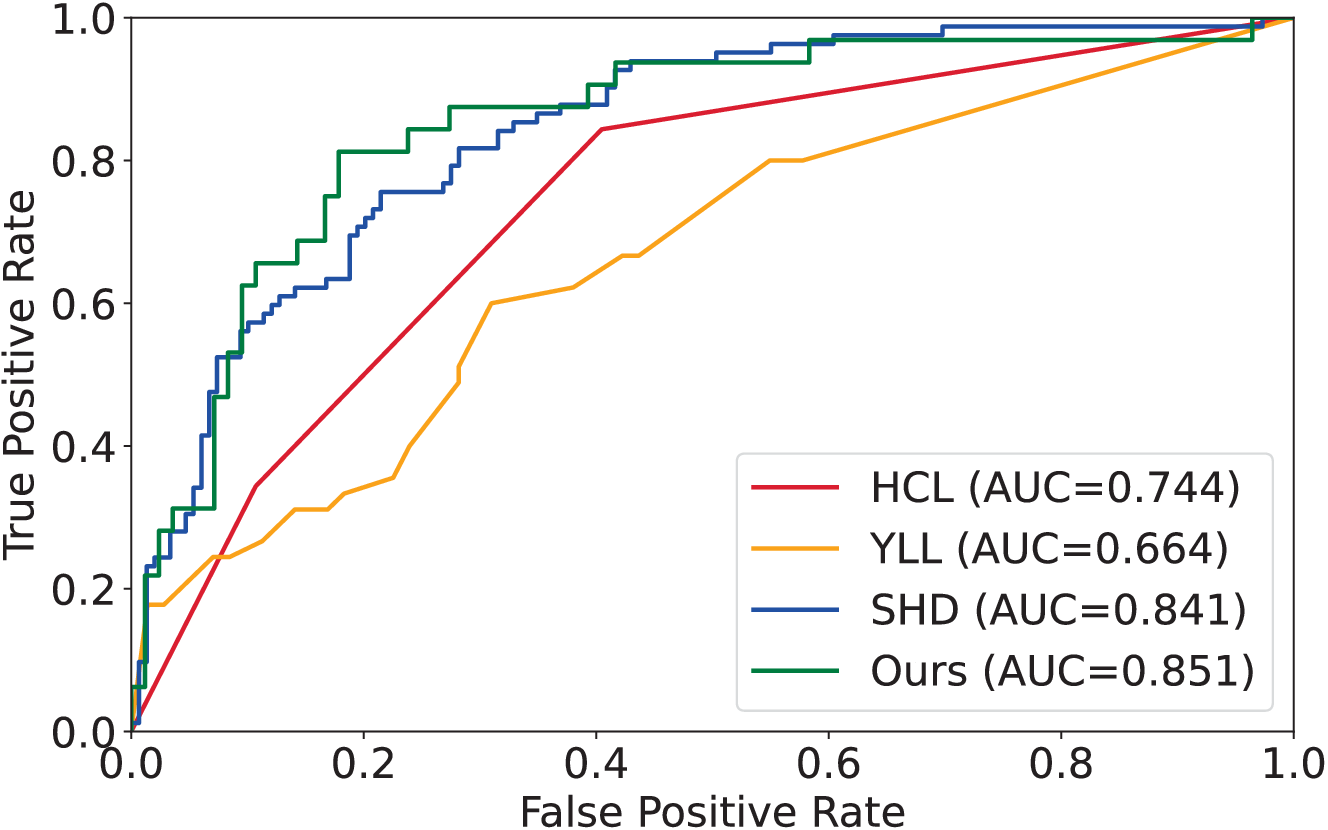}}

  \subfigure[]{\includegraphics[width=0.45\linewidth,height=0.278\linewidth]{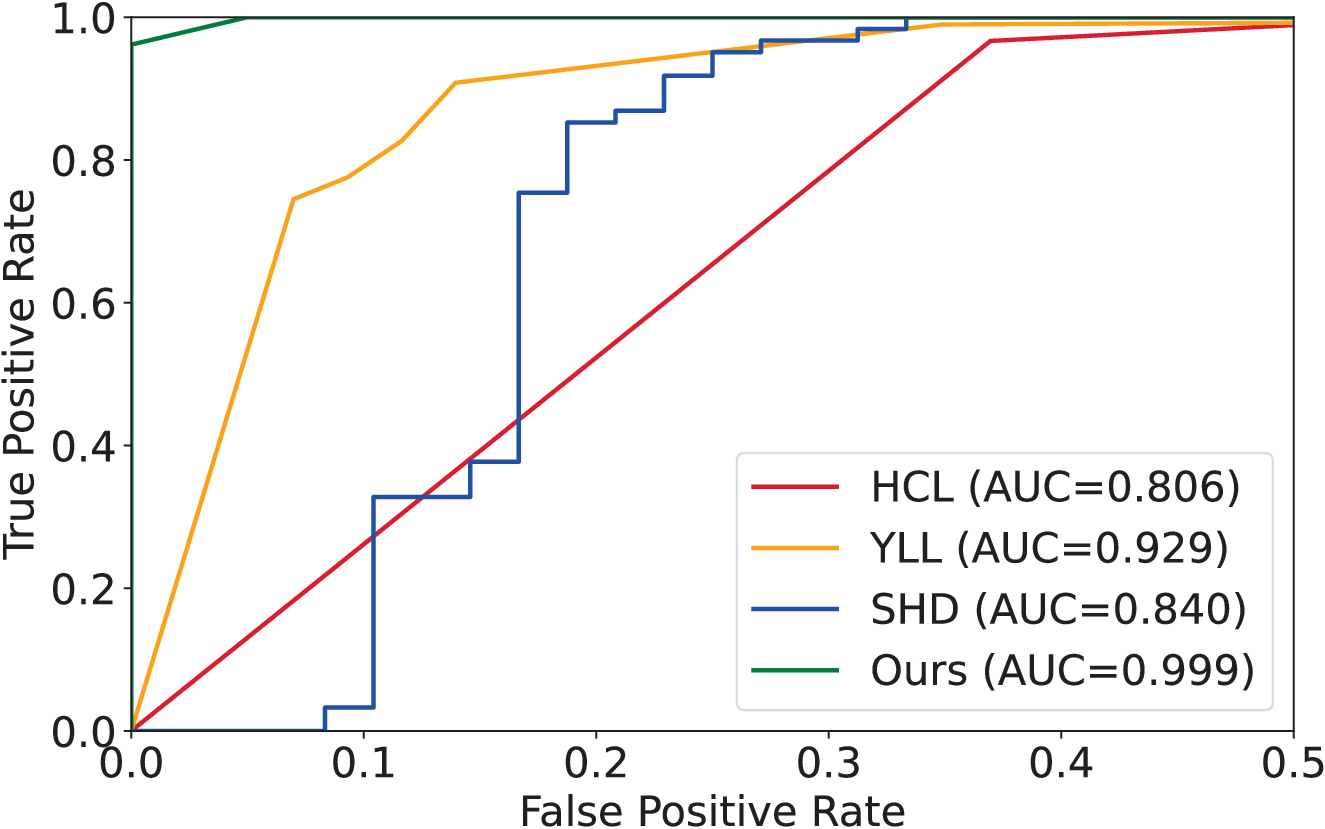}}
  \subfigure[]{\includegraphics[width=0.45\linewidth,height=0.278\linewidth]{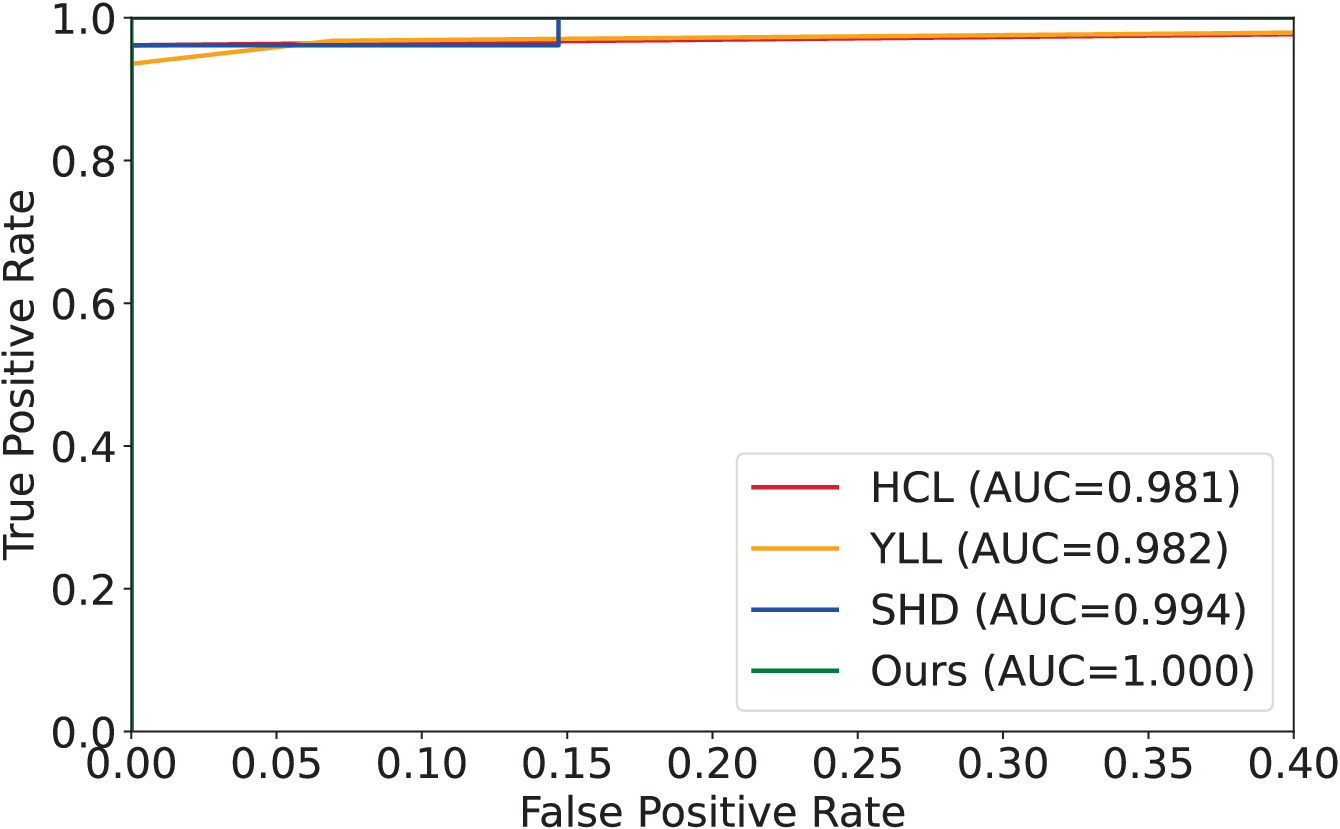}}

  \caption{ROC and AUC on eight datasets. (a) Waveform dataset; (b) Spambase dataset; (c)Sonar dataset; (d) Wdbc dataset; (e) Clean dataset; (f) Pima dataset; (g) Ionosphere; (h) Divorce dataset.}
\label{fig1}
\end{figure}

In summary, from Table \ref{experiment}, the proposed method achieves good results on a variety of datasets, especially on Precision and Specificity, which implies that the proposed method is likely to be a powerful tool for binary classification tasks. However, it should be noted that on specific datasets, such as Pima, the Recall of the proposed method is relatively low, which may be an area for further research and optimization.
Similiary, as can be seen from Fig. \ref{fig1}, the proposed algorithm performs very well on all datasets, and its AUC values are usually the highest, indicating that its performance on the classification task is very robust and effective. The performance of the other three methods also varies, but overall the proposed method is the clear winner in these comparisons.

\subsection{Statistical Analysis}

The purpose of statistically analyzing the classification results of the above four methods on eight datasets is to further demonstrate the superiority of the method proposed in this paper over the other methods. In addition, a difference test can be performed to detect whether there is a meaningful difference between the four methods mentioned above. Therefore, Friedman test is used for statistical analysis \cite{ref60}. It is a powerful strategy for conducting statistical analysis, and used to measure the significant differences between the four methods in the comparative experiments. In this paper, Friedman test relies on a significance level of $\alpha$ = 0.05 and is used to average the G-means obtained from all eight UCI dataset species.

In order to effectively account for the efficiency of each method, Friedman test also ranks each method retrogradely according to its effectiveness on each dataset. The lower the ranking score for a method, the better its performance and the more efficient it is.
\begin{figure}[htbp]
\centering
\includegraphics[width=1\linewidth]{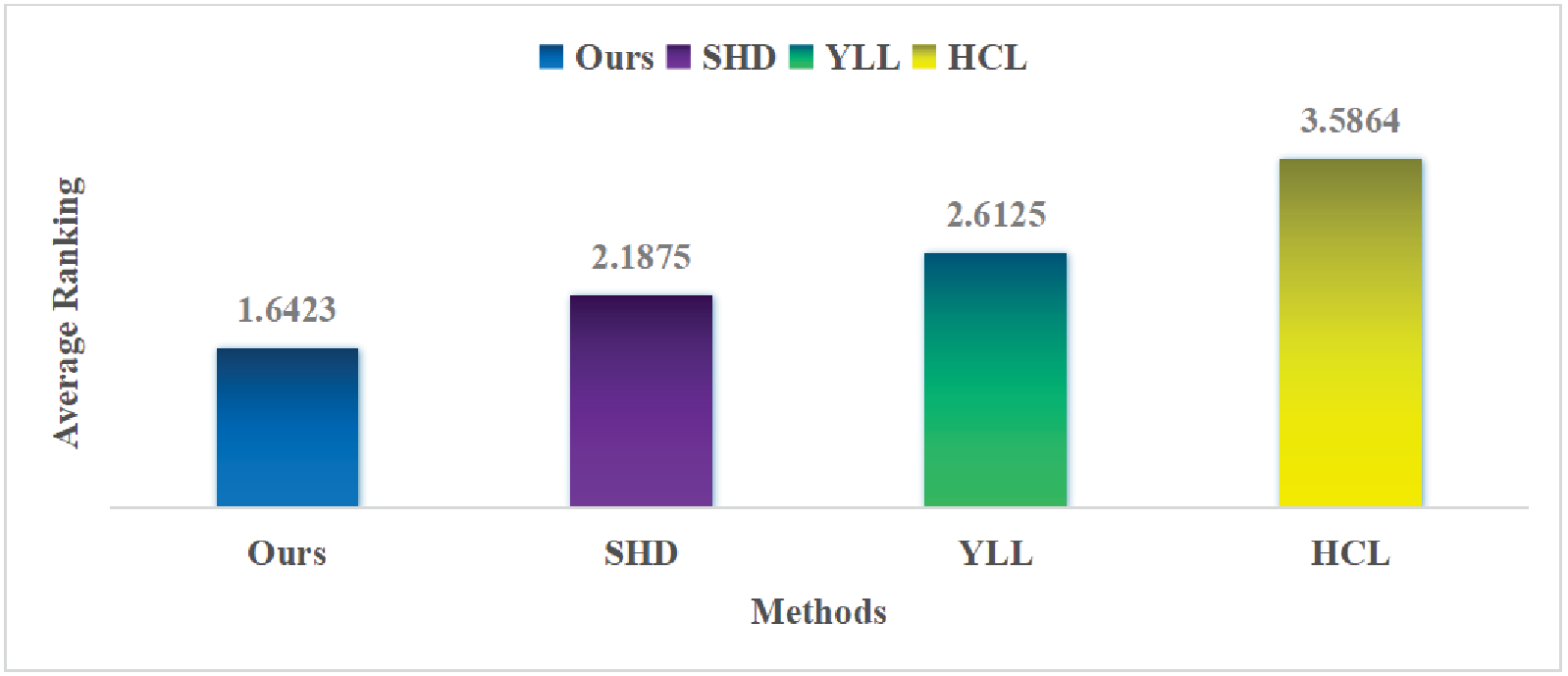}
\caption{Summary of the average rank of the four binary classification methods.}
\label{test}
\end{figure}

From Fig. \ref{test}, it can be seen that the proposed algorithm has the highest performance with an average rank of 1.6423. SHD has the second highest performance with an average rank of 2.1785, and YLL as well as HCL are in the third and fourth places, with an average rank of 2.6125 and 3.5864, respectively.

\subsection{Ablation Experiment}

In order to better realize feature selection and get a better subset of features, we construct an iterative feature selection framework based on fuzzy correlation family (FCF) in combination with the biclustering algorithm. In addition, when extracting classification rules based on the obtained biclusters, we introduce the fuzzy membership function in order to retain the complete classification information contained in the biclusters. Thus, the exact classification rules extracted from the biclusters are transformed into fuzzy rules (FR) to further improve the performance of the classification rules.

\begin{figure}[htbp]
\centering
\includegraphics[width=1\linewidth]{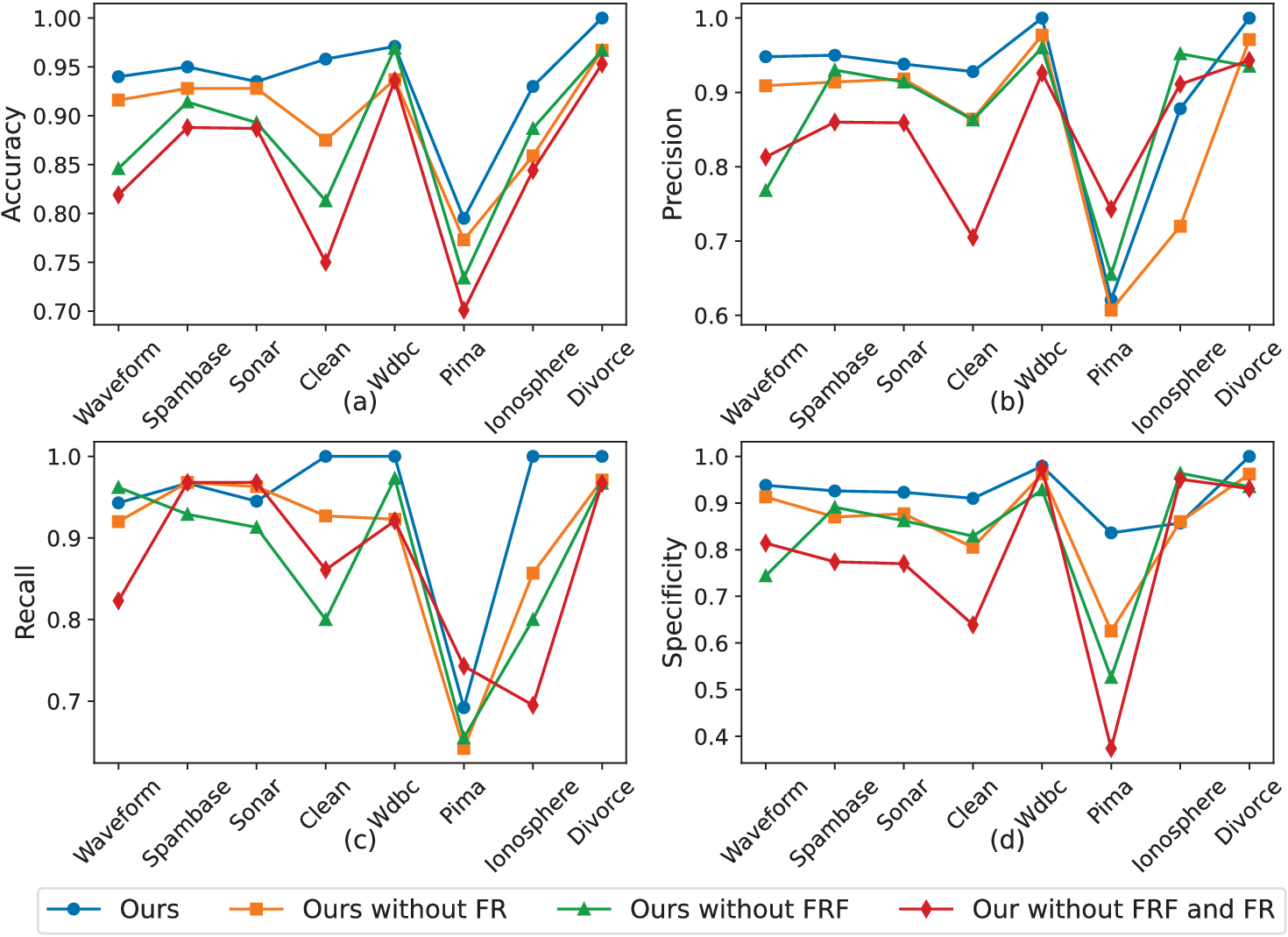}
\caption{Ablation experimental results. (a) the accuracy for each algorithm; (b) the precision for each algorithm; (c) the recall for each algorithm; (d) the specificity for each algorithm.}
\label{ablation}
\end{figure}

In order to verify the effects of both FCF and FR, we conduct ablation experiments on eight UCI datasets and observe the performance of the four algorithms on the four metrics, Accuracy, Precision, Recall, and Specificity. In Table \ref{tablex}, we show four different algorithms and their average values for the four metrics experimented on eight datasets. In Fig. \ref{ablation}, we show the values of each metric on each dataset for each of the four algorithms separately.

\begin{table}[htbp]
\centering
\caption{\textsc{Algorithms after combination}}
\tiny
\label{tablex}
\resizebox{1\linewidth}{!}{
\begin{tabular}{cccccc}
\toprule
FCF&FR&Accuracy&Precision&Recall&Specificity\\
\midrule
\multicolumn{1}{c}{\checkmark} &\multicolumn{1}{c}{\checkmark} &\multicolumn{1}{c}{0.935} &\multicolumn{1}{c}{0.883} &\multicolumn{1}{c}{0.943} &\multicolumn{1}{c}{0.921}\\
\multicolumn{1}{c}{\checkmark} &\multicolumn{1}{c}{$\times$} &\multicolumn{1}{c}{0.898}&\multicolumn{1}{c}{0.860}&\multicolumn{1}{c}{0.896}&\multicolumn{1}{c}{0.859}\\
\multicolumn{1}{c}{$\times$} &\multicolumn{1}{c}{\checkmark} &\multicolumn{1}{c}{0.878}&\multicolumn{1}{c}{0.835} &\multicolumn{1}{c}{0.869} &\multicolumn{1}{c}{0.820}\\
\multicolumn{1}{c}{$\times$} &\multicolumn{1}{c}{$\times$} &\multicolumn{1}{c}{0.847} &\multicolumn{1}{c}{0.808} &\multicolumn{1}{c}{0.868} &\multicolumn{1}{c}{0.761}\\
\bottomrule
\end{tabular}
}
\end{table}

As can be seen from Table \ref{tablex} and Fig. \ref{ablation}, removing either FCF or FR module from the proposed algorithm results in a decrease in Accuracy, Recall, Precision, and Specificity. Overall, both FR and FCF components play a key role in the algorithm performance and their ablation leads to performance degradation. FCF seems to have a greater impact on the performance of the algorithm than FR, especially in terms of Recall.

\section{Conclusion}
Traditional classification algorithms have two problems: 1) the feature selection process and classification process are separated, resulting in incomplete feature extraction; 2) when mining classification rules, the ambiguity of rule is ignored, resulting in the loss of classification information. In order to solve these problems, we propose a fuzzy rule-based binary classification algorithm with an iterative feature selection framework. In it, we use feature selection based on fuzzy correlation family and a heuristic biclustering algorithm to build the iterative feature selection framework. We continuously select features through feedback of bicluster evaluation results so that features and biclusters can be optimized at the same time. In addition, by introducing a rule membership function when extracting rules, we increase the fuzziness of rules and successfully build a strong classifier through AdaBoost.

Of course, the proposed algorithm has several limitations. If the experiments can be conducted on newer as well as more diverse datasets, it can be more convincing. In addition, the biclustering methods used in this paper are traditional biclustering based on heuristic algorithms, without using the most advanced current methods, such as evolutionary biclustering. In future research, we will try to build a more complex and effective framework for feature selection and further optimize the algorithm to get a better feature subset. In addition, we will attempt to improve the algorithm in the future and combine it with methods such as neural networks, so as to apply the algorithm to more fields such as image recognition.

\vspace{5pt}
\begin{IEEEbiography}[{\includegraphics[width=1in,height=1.25in,clip,keepaspectratio]{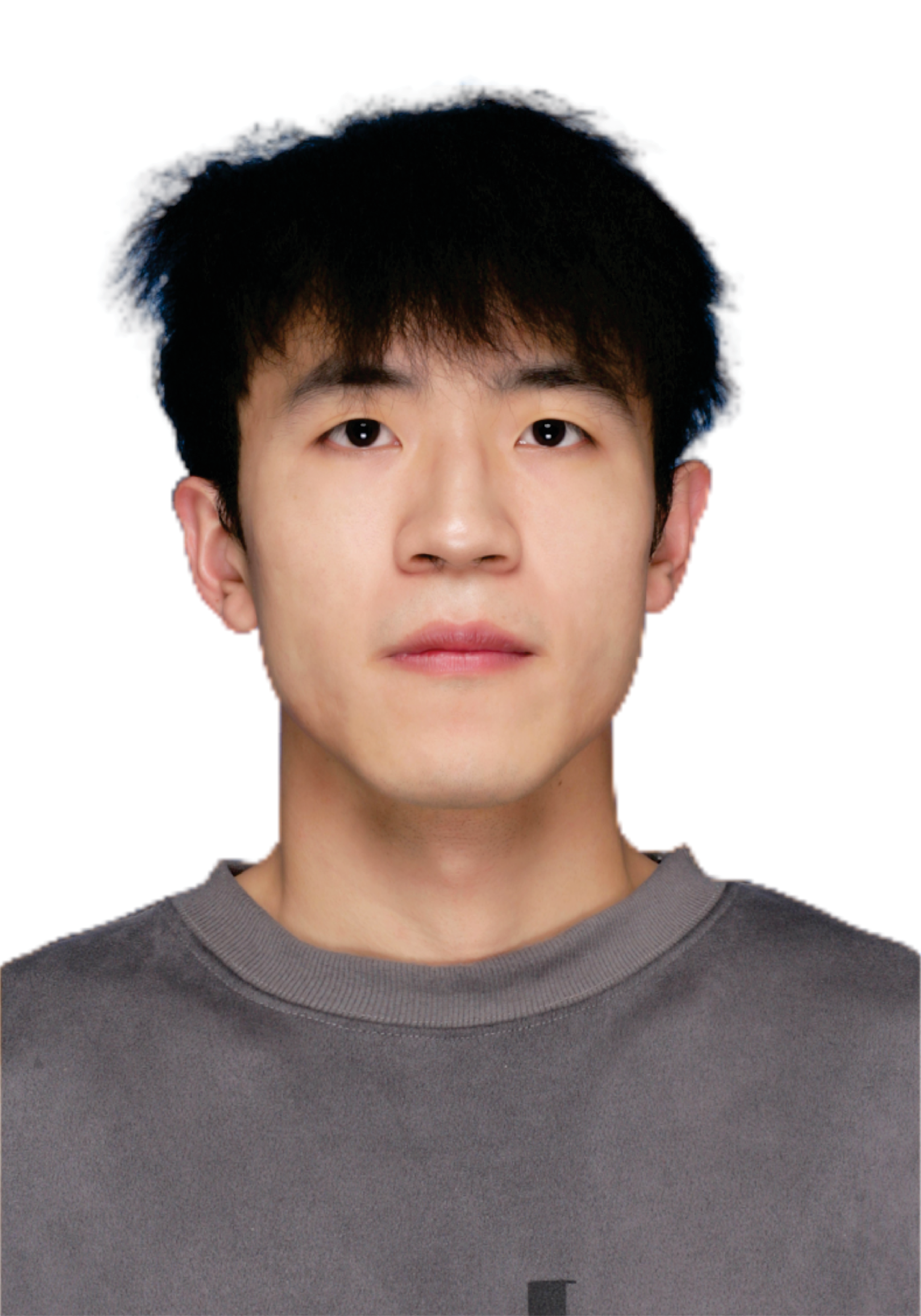}}]{Haoning Li}
received the B.S. degree and the M.S. degree in computer science and technology from Northwestern Polytechnical University. He is currently working toward the Ph.D. degree in artificial intelligence with Northwestern Polytechnical University, Xi'an, China.
\end{IEEEbiography}

\begin{IEEEbiography}[{\includegraphics[width=1in,height=1.25in,clip,keepaspectratio]{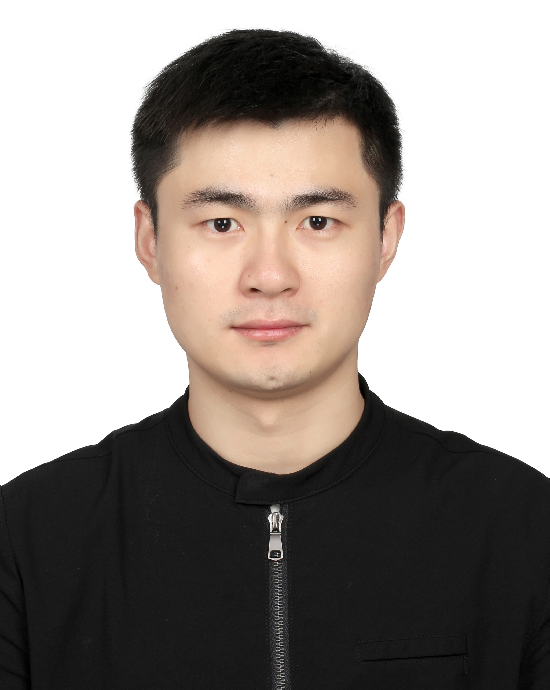}}]{Cong Wang}
(Member, IEEE) received the B.S. degree in automation and the M.S. degree in mathematics from Hohai University, Nanjing, China, in 2014 and 2017, respectively. He received the Ph.D. degree in mechatronic engineering from Xidian University, Xi'an, China in 2021.

He is now an associate professor in School of Artificial Intelligence, OPtics and ElectroNics (iOPEN), Northwestern Polytechnical University, Xi'an, China. He was a Visiting Ph.D. Student at the Department of Electrical and Computer Engineering, University of Alberta, Edmonton, AB, Canada, and the Department of Electrical and Computer Engineering, National University of Singapore, Singapore. He was also a Research Assistant at the School of Computer Science and Engineering, Nanyang Technological University, Singapore. His current research interests include wavelet analysis and its applications, fuzzy theory, granular computing, as well as image processing. 
\end{IEEEbiography}

\begin{IEEEbiography}[{\includegraphics[width=1in,height=1.25in,clip,keepaspectratio]{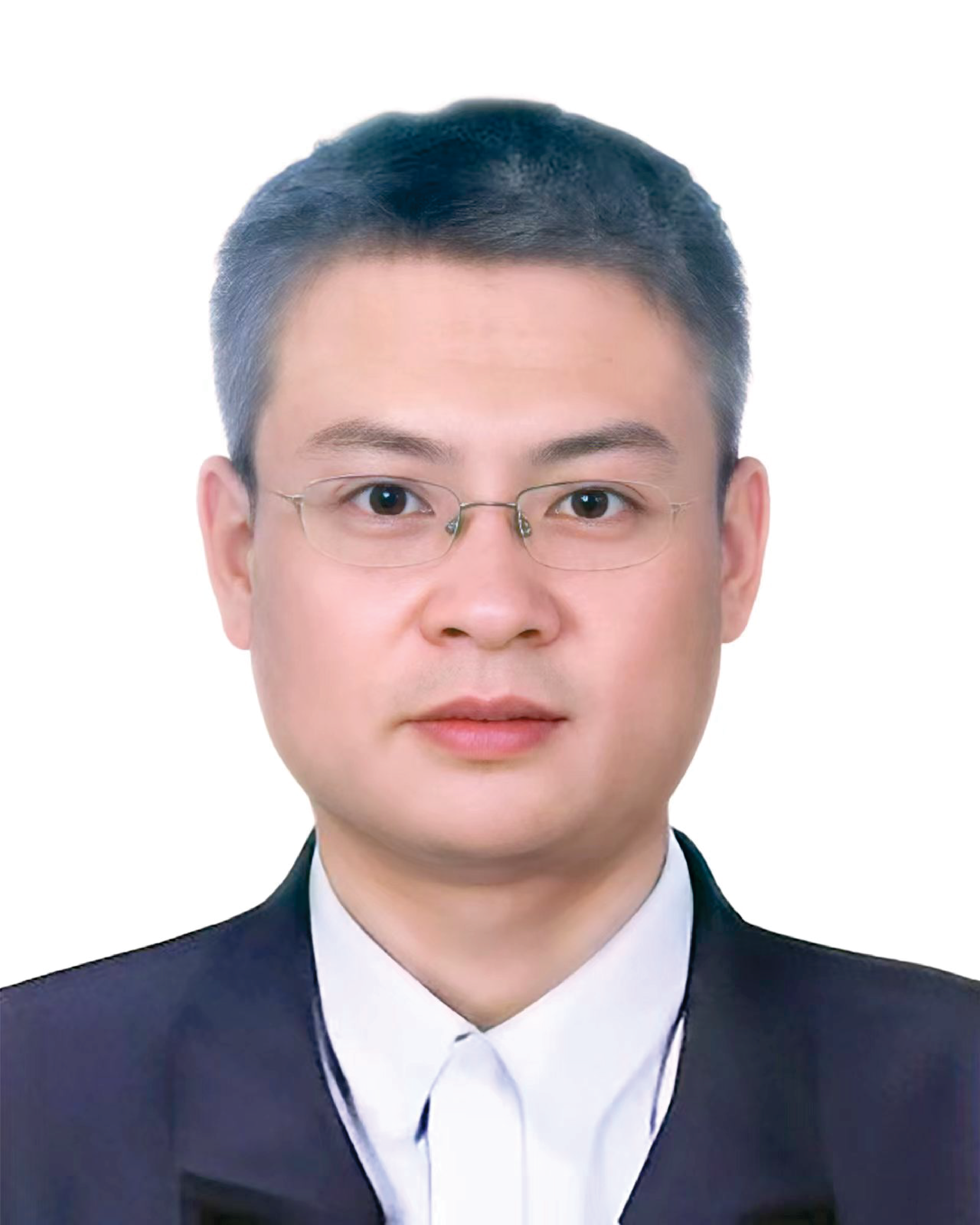}}]{Qinghua Huang}
received the Ph.D. degree in biomedical engineering from the Hong Kong Polytechnic University, Hong Kong, in 2007. Now he is a full professor in School of Artificial Intelligence, Optics and Electronics (iOPEN), Northwestern Polytechnical University, China. His research interests include multi-dimensional u1trasonic imaging, medical image analysis, machine learning for medical data, and intelligent computation for various applications. 
\end{IEEEbiography}

\clearpage
\vfill

\end{document}